\documentclass[conference]{IEEEtran}
\IEEEoverridecommandlockouts
\usepackage{subcaption} 
\usepackage{colortbl}
\usepackage{booktabs}
\usepackage{enumitem}
\usepackage[normalem]{ulem}
\usepackage[ruled,vlined]{algorithm2e}
\usepackage{tabularx}
\newcolumntype{Y}{>{\centering\arraybackslash}X}
\usepackage{cite}
\usepackage{amsmath,amssymb,amsfonts}
\usepackage{algorithmic}
\usepackage{graphicx}
\usepackage{textcomp}
\usepackage{xcolor}
\usepackage{fontawesome5}

\definecolor{orcidlogocol}{HTML}{A6CE39}
\newcommand{\orcid}[1]{%
  \textsuperscript{%
    \edef\@orcidlink{https://orcid.org/#1}%
    \pdfannot width 6pt height 6pt depth 0pt {
      /Subtype /Link
      /Border [0 0 0]
      /A << /S /URI /URI (\@orcidlink) >>
    }%
    \textcolor{orcidlogocol}{\normalfont\faOrcid}%
  }%
}

\newtheorem{myDef}{Definition}
\def\BibTeX{{\rm B\kern-.05em{\sc i\kern-.025em b}\kern-.08em
    T\kern-.1667em\lower.7ex\hbox{E}\kern-.125emX}}
\begin{document}

\title{AHEAD: Advancing Multi-Class Label Aggregation with Interpretable Cross-Annotator Modeling
}
\author{
\IEEEauthorblockN{
Ju Chen\textsuperscript{1}, Sijia Xu\textsuperscript{2}, Jun Feng\textsuperscript{1}, Zhiqiang Gao\textsuperscript{3}, Zhengyi Yang\textsuperscript{4,\textasteriskcentered}
}
\IEEEauthorblockA{
\textsuperscript{1}Hohai University, China \textsuperscript{2}University of New South Wales, Australia\\
\textsuperscript{3}Southeast University, China \textsuperscript{4}The University of Sydney, Australia\\
JuJuChen2026@outlook.com, sijia.xu@unsw.edu.au, fengjun@hhu.edu.cn, zqgao@seu.edu.cn\\
zhengyi.yang@sydney.edu.au
}

\thanks{\textasteriskcentered~Corresponding author.}
\thanks{Preprint. Under review.}
}

\maketitle

\begin{abstract}
Crowdsourced labeling provides valuable labeled data for domains across natural language processing, computer vision, and video. Label aggregation aims to infer latent true labels from noisy and biased annotations, with the key lying in annotator reliability estimation. Despite promising progress, existing approaches struggle with one real-world bottleneck: most individual annotators label only a small subset of tasks, making accurate annotator estimation highly intractable. In this paper, we focus on the considerably more challenging multi-class label aggregation and propose \textbf{AHEAD} (cross-\textbf{A}nnotator learning and \textbf{H}igh-confid\textbf{E}nce \textbf{A}nnotator-guide\textbf{D} label aggregation), a cross-annotator learning framework that advances annotator reliability estimation by leveraging the population-level data. Specifically, AHEAD first learns high-dimensional cross-annotator contexts via a graph neural network, deriving multi-view, complementary annotator embeddings by aggregating individual-level annotator features with contextual information. These embeddings are then decoded into interpretable annotator-specific confusion matrices to fit the observed labels. We formulate a composite objective incorporating high-confidence annotators to alleviate the unsupervised training issues faced by prior models. Experiments on 10 real-world datasets spanning NLP, CV, Video, and Audio show that AHEAD substantially improves label accuracy, increasing average accuracy from 68.75\% to 73.23\%, with gains of up to 14.9\% in the best case. Meanwhile, scalability experiments on the largest dataset further demonstrate the overall superiority of our method.
\end{abstract}


\section{Introduction}
By harnessing crowd intelligence, crowdsourced labeling continuously injects massive, valuable labeled data into areas including natural language processing, computer vision, and video \cite{li2020,modaresnezhad2020,Das2023state,Sunetal24}, expanding both data breadth and knowledge depth \cite{Zhengetal22,Qiuetal24,Zhangetal224}. Meanwhile, it has become indispensable for evaluating and fact-checking large language models \cite{Otanietal23,Li24,Deetal25,Gienappetal25}. Considering the varying reliability of different annotators \cite{Chaietal19}, crowdsourcing platforms typically assign each task to multiple annotators and apply label aggregation to infer the latent true label (truth) \cite{Lietal2016,Zhengetal2017}. Majority Voting (MV), which treats the label provided by most annotators as the truth, is the most naive method but is prone to bias as it fails to account for annotator reliability \cite{Hornetal18}. Consequently, a wide spectrum of fine-grained approaches has emerged to jointly infer annotator reliability and latent truths. In this paper, we focus on the \textbf{multi-class label aggregation}, which is considerably common and more challenging than the binary cases due to complex inter-class confusion and annotator noise \cite{Chu21a,Yangetal24,Zhangetal25b}.

Depending on how annotators are modeled, existing approaches can be grouped into \textit{single-parameter-based} \cite{Zhouetal12,Lietal2014,Lietal14,Lietal2019b}, \textit{confusion-matrix-based} \cite{DawidSkene1979,KimGhahramani2012,Zhangetal16,Zhangetal24,Chenetal25}, and \textit{representation-learning-based} \cite{Yinetal17,Liuetal21,Wuetal2023a,Wuetal2023b,Liuetal24,liu2026crowdfm} methods. By modeling annotator reliability with a scalar accuracy shared across all classes, single-parameter-based methods \cite{Whitehilletal2009,Lietal2014,Lietal2019b,Caoetal2020} impose a strong homogeneity assumption over diverse classes. Therefore, confusion-matrix-based and representation-learning-based methods currently dominate due to their extensive characterization of annotators.

Confusion-matrix-based methods, which model each annotator with a confusion matrix, are theoretically well-founded for capturing the label generation process \cite{Lietal2019b}. The DS model \cite{DawidSkene1979} is the first to introduce annotator-specific confusion matrices to characterize annotator reliability. Subsequent extensions incorporate more expressive priors: IBCC \cite{KimGhahramani2012} adopts a Dirichlet prior over confusion matrices, FGBCC \cite{Chenetal25} develops a Softmax-Gaussian prior, and EBCC \cite{Lietal2019a} further extends IBCC \cite{KimGhahramani2012} by partitioning tasks with the same truth into subtypes. More recently, Zhang et al. \cite{Zhangetal24} propose a coupled method to jointly refine confusion matrices.
\begin{figure}[h]
    \centering
    \begin{subfigure}[b]{0.32\linewidth}
        \centering
        \includegraphics[width=\textwidth]{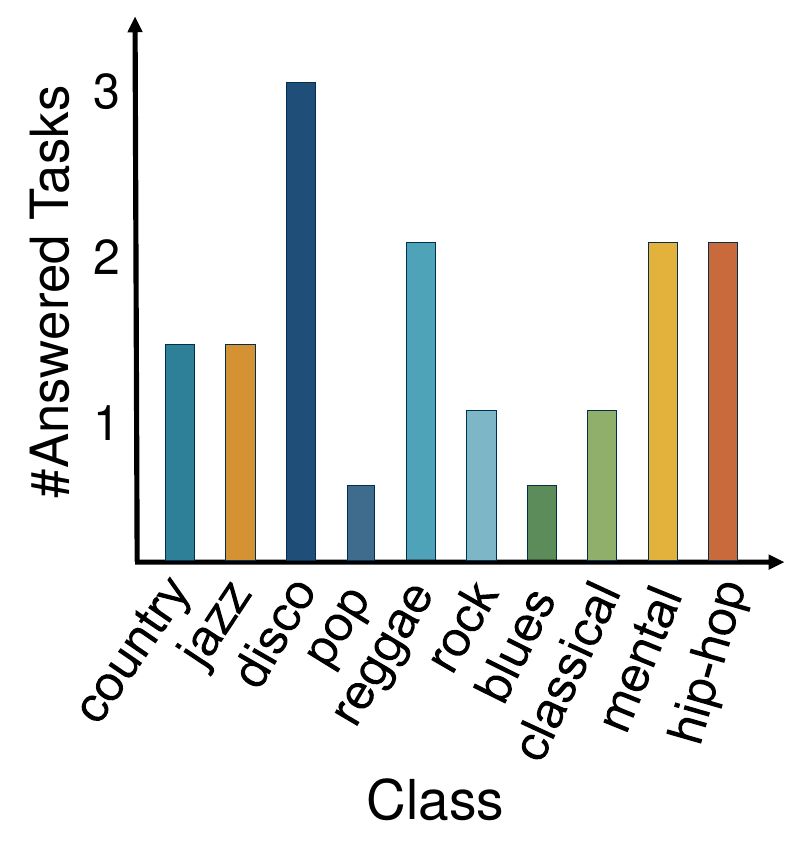}
        \caption{}
        \label{val5}
    \end{subfigure}
    \hfill
    \begin{subfigure}[b]{0.66\linewidth}
        \centering
        \includegraphics[width=\textwidth]{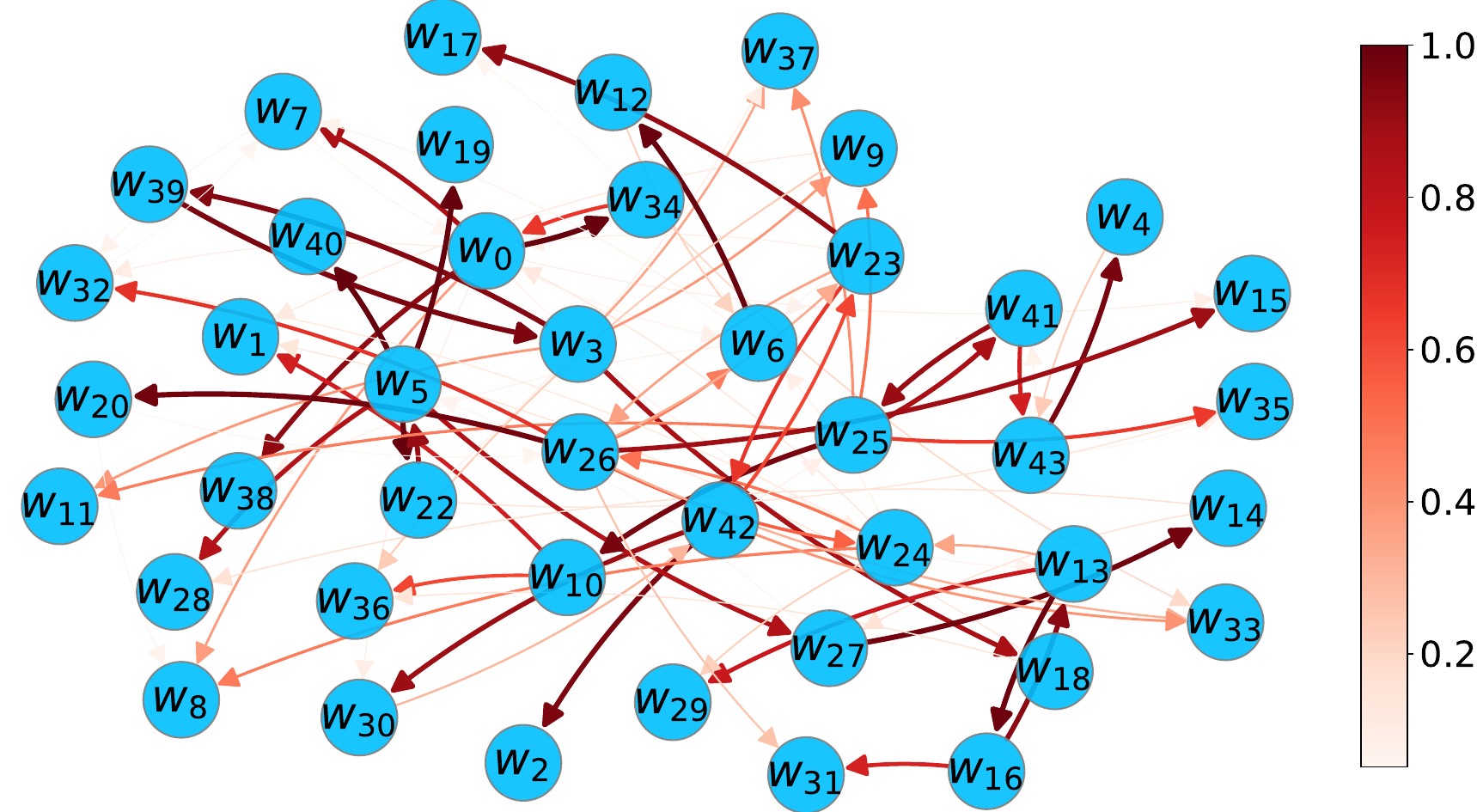}
        \caption{}
        \label{ms}
    \end{subfigure}
    \caption{(a) Average class-wise annotation counts for annotators with the median number of annotations in the MS dataset. (b) Cross-annotator correlations learned on the MS dataset, where $w_{i}$ denotes the $i$-th annotator.}
    \label{intro}
\end{figure}

Empowered by their capacity to learn complex multi-dimensional embeddings, representation-learning-based methods have recently proliferated. Wu et al. \cite{Wuetal2023b} propose a hierarchical graph autoencoder to model multi-view interactions within a task-annotator graph, complemented by an initialization strategy that iteratively refines node embeddings and annotator reliability. Liu et al. \cite{Liuetal24} apply graph contrastive learning on an annotator-task bipartite graph, generating augmented views via edge perturbation to promote view invariance. Recently, Liu et al. \cite{liu2026crowdfm} pretrain a bipartite graph neural network on a series of simulated labeling scenarios to learn diverse annotation patterns and generalize across datasets without dataset-specific training.

Although confusion-matrix-based and representation-learning-based methods are well-grounded and highly expressive, they struggle with a fundamental real-world bottleneck: \textbf{most individual annotators label only a small subset of tasks}, a ubiquitous phenomenon in real-world applications \cite{Ibrahietal23,Zhangetal24} that makes accurate annotator estimation highly intractable. For instance, Figure \ref{intro}(a) illustrates the average class-wise annotation counts for annotators with the median number of annotations in the real-world dataset MS \cite{RodriguesPR13}, revealing substantial data sparsity in overall annotation volume, and the \textbf{class-level sparsity} and \textbf{imbalance} are even more pronounced, as most annotators cover only a few samples for many classes. Moreover, current methods mainly build upon low-density annotator-task topology, making the inference unreliable. In contrast, we argue that cross-annotator correlations are more dense and reveal multi-view, complementary annotator reliability. For instance, Figure \ref{intro}(b) depicts the dense cross-annotator correlations that we learn from MS, with the color bar indicating the correlation degree. These correlations offer us population-level information to derive multi-view, complementary assessments of individual annotators.

To this end, this work aims to achieve accurate multi-class label aggregation under sparse data. Intuitively, confusion matrices offer an interpretable and theoretically grounded characterization of the label generation process \cite{Lietal2019b}, while representation learning excels at capturing latent nonlinear correlations that are difficult to model explicitly. Therefore, we first propose to adopt representation learning to model high-dimensional cross-annotator correlations and derive multi-view, complementary annotator embeddings from the population. Then, we decode these embeddings into interpretable, annotator-specific confusion matrices to fit the observed labels. We thereby advance annotator reliability estimation by leveraging the population data. Since we operate without ground-truth supervision, two critical issues must be addressed:
\begin{itemize}[leftmargin=*]
\item Learning expressive annotator embeddings that map to valid confusion matrices. It is non-trivial to derive multi-view, complementary annotator embeddings and project them into valid annotator confusion matrices in a principled manner.
\item Establishing a robust optimization objective. Prior approaches typically rely on Majority Voting (MV)-derived pseudo-labels for model training. However, MV-derived labels are notoriously noisy, and training directly on such signals can deteriorate the performance.
\end{itemize}

Building upon the above insights, we propose \textbf{AHEAD} (cross-\textbf{A}nnotator learning and \textbf{H}igh-confid\textbf{E}nce \textbf{A}nnotator-guide\textbf{D} label aggregation). Specifically,

(1) to address the first issue, AHEAD \uline{explicitly models high-dimensional cross-annotator learning} over their contextual topology and learns multi-view, complementary annotator embeddings by aggregating individual-level annotator features with their contextual information via a graph neural network. These embeddings are then decoded into \uline{interpretable annotator-specific confusion matrices} to fit the observed labels. We introduce a negative log-likelihood loss using the decoded confusion matrices to maximize the likelihood of the observations, ensuring expressive annotator embeddings and valid, statistically sound annotator confusion matrices.

(2) to address the second issue, we formulate a composite objective that integrates \uline{likelihood maximization}, \uline{structural regularization}, and \uline{directional guidance}. Specifically, we identify annotators whose task counts fall within a top percentage as high-confidence annotators and leverage their empirical confusion matrices as critical supervisory signals by minimizing the discrepancy between the predicted and empirical confusion matrices. This mechanism is motivated by the Law of Large Numbers \cite{Durrett2010}, which suggests that an empirical estimate becomes increasingly accurate as the sample size grows. Therefore, although Majority Voting (MV) can be noisy, the empirical confusion matrices of high-confidence annotators are statistically trustworthy and thereby alleviate the unsupervised training issues faced by prior models.

Extensive experiments across 10 real-world datasets spanning five domains demonstrate the effectiveness and efficiency of AHEAD, and scalability experiments on the largest dataset demonstrate the comprehensive superiority of our method.

In summary, the contributions of this study are as follows:
\begin{itemize}[leftmargin=*]
    \item We propose cross-annotator learning to advance individual annotator reliability estimation by leveraging population-level data and unifying it with interpretable confusion matrices to fit the observed labels, substantially improving the accuracy of multi-class label aggregation.
    \item We identify high-confidence annotators and formulate a composite objective incorporating likelihood maximization, structural regularization, and directional guidance to alleviate the unsupervised training issues faced by prior models.
    \item Extensive experiments on 10 real-world datasets spanning NLP, CV, Video, and Audio demonstrate that we achieve a 14.9\% accuracy improvement in the best case and boost the average accuracy from 68.75\% to 73.23\%. Experiments on the largest dataset confirm the scalability of our method.
\end{itemize}
\section{Related Works} \label{related-work}
This paper focuses on crowdsourced label aggregation that infers truths and annotator reliability solely from (task, annotator, label) triples. Notably, there are also related works for scenarios with rich task features available \cite{TannoSSAS19,Lietal21,IbrahimN023,Tanetal24,Zhangetal25a} or mobile crowdsourcing where annotators perform sensing tasks via mobile devices in specific environments \cite{Wuetal22,Baietal25,Liuetal25}, which are outside the scope of our detailed discussion.

Depending on how annotators are modeled, existing studies can be divided into single-parameter-based \cite{Demartinietal2012,Bonald17,Lietal2019b,Caoetal2020}, confusion-matrix-based \cite{KimGhahramani2012,Lietal2019a,Songetal2021,Zhangetal24,Chenetal25}, and representation-learning-based \cite{Yinetal17,Lyuetal21,Wuetal2023a,Wuetal2023b,Liuetal24} methods.
\subsection{Single-Parameter-Based Methods}
Single-parameter-based methods assume that each annotator maintains a scalar accuracy across all classes. The GLAD model \cite{Whitehilletal2009} incorporates task difficulty and annotator reliability to characterize the log odds of each label being true and infers the parameters using the expectation-maximization framework. The CATD model \cite{Lietal2014}  models annotator errors as normally distributed and formulates an objective to minimize the weighted sum of error variances, with annotator reliability updated via the upper bound of the variance's confidence interval. Aydin et al. \cite{Aydinetal14} also minimize a weighted sum of annotators' errors, constraining annotator reliability to be non-negative and summing to one, and infer using block coordinate descent techniques. Bonald and Combes \cite{Bonald17} derive a lower bound on the minimax estimation error and propose triangular annotator estimation to ensure non-asymptotic performance. BWA \cite{Lietal2019b} and OKELE \cite{Caoetal2020} are both generative Bayesian models. The BWA model \cite{Lietal2019b} models annotator reliability with a Gamma distribution and updates parameters via the expectation-maximization framework. The OKELE model \cite{Caoetal2020} models annotator reliability with a scaled inverse chi-squared distribution and updates parameters via a gradient method.

Generally, single-parameter-based methods are computationally efficient, but their strong homogeneous annotator reliability across classes makes them suboptimal in practice.
\subsection{Confusion-Matrix-Based Methods}
Confusion-matrix-based methods exploit annotator-specific confusion matrices to distinguish annotators' varying reliability across classes. Assuming that annotators are independent, the DS model \cite{DawidSkene1979} is the first to model each annotator using a confusion matrix within a Bayesian generative framework, with parameters updated through maximum likelihood estimation. The IBCC model \cite{KimGhahramani2012} and FGBCC model \cite{Chenetal25} extend DS by introducing Dirichlet and Softmax-Gaussian priors for the rows of confusion matrices, respectively. Specifically, IBCC \cite{KimGhahramani2012} utilizes Gibbs sampling for inference, whereas FGBCC \cite{Chenetal25} adopts a variational inference approach. The CBCC model \cite{Venanzietal2014} posits the existence of latent annotator communities, assuming that each annotator's confusion matrix aligns closely with that of their respective community. The DBCC model \cite{Venanzietal2014} exploits a Markov network to model the dependence between annotators and updates the parameters via a Metropolis sampling method. The EBCC model \cite{Lietal2019a} extends IBCC \cite{KimGhahramani2012} by dividing tasks with the same truth into subtypes. Chu et al. \cite{Chu21b} employ both global and individual confusion matrices to characterize annotator behavior. Specifically, they introduce a Bernoulli variable to model the label generation process, conditioned on task difficulty and annotator reliability. Ibrahim et al. \cite{Ibrahimetal19}, \cite{Ibrahim21} estimate annotator reliability through a matrix factorization that leverages annotator co-occurrence.

Confusion-matrix-based methods are theoretically well-founded through explicit modeling of the label generation process. However, ubiquitous data sparsity make full confusion matrix estimation highly unstable.
\begin{figure*}
\includegraphics[width=\textwidth]{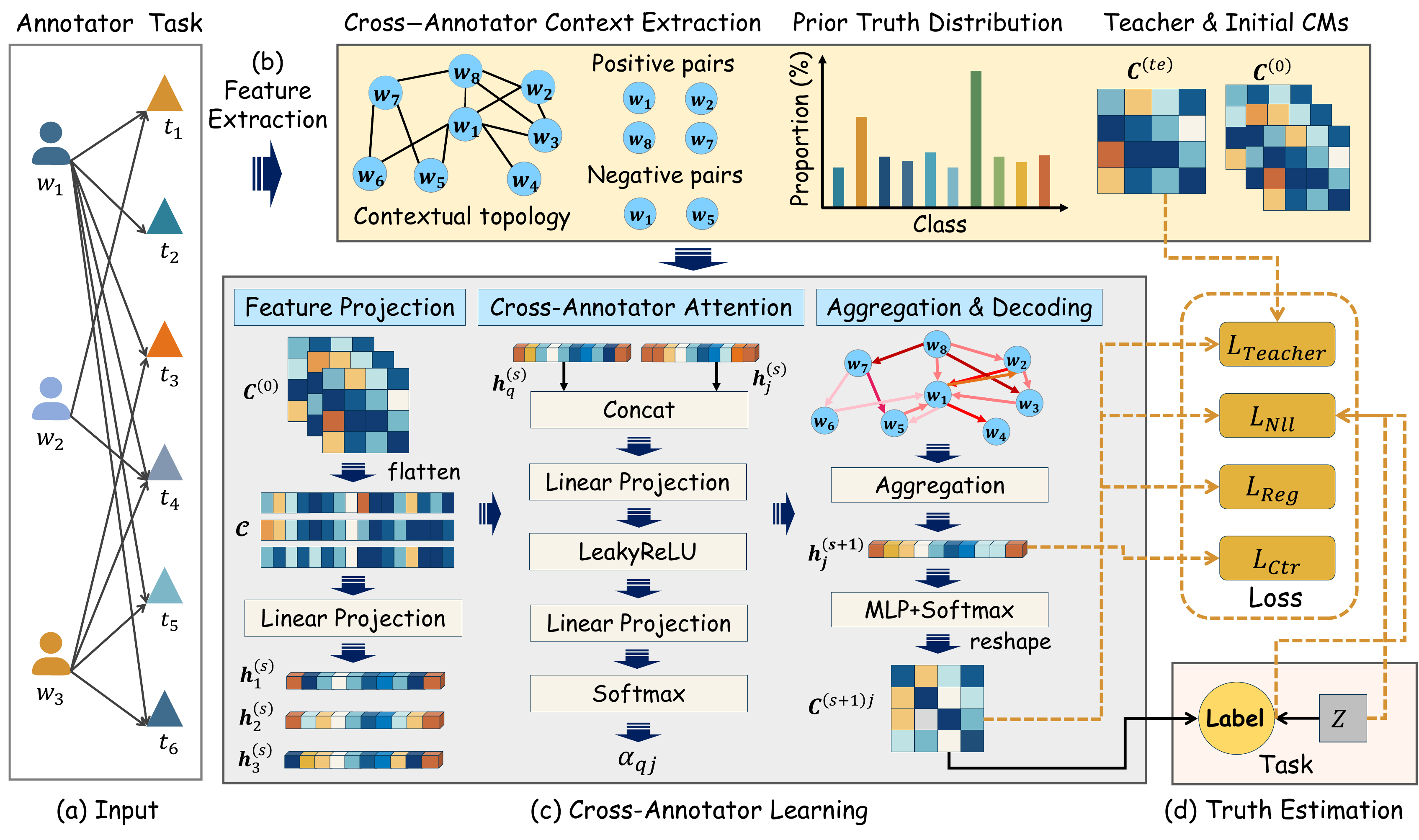}
\caption{Overview of the proposed AHEAD, which consists of four components: (a) Input; (b) Feature extraction, which derives cross-annotator contexts, teacher and initial confusion matrices, and prior truth distribution; (c) Cross-annotator learning, where the graph attention layer learns cross-annotator correlations, the encoder layer derive multi-view, complementary annotator embeddings, and the decoder layer predict annotator-specific confusion matrices; (d) Truth estimation, which updates posterior truth probabilities based on the updated parameters.} \label{framework}
\end{figure*}
\subsection{Representation-Learning-Based Methods}
Representation-learning-based methods embed annotators, tasks, and labels into continuous spaces to capture latent relationships. Yin et al. \cite{Yinetal17} propose an unsupervised autoencoder framework that integrates a classifier and a reconstructor: the former infers truths from the input, while the latter reconstructs the input from these inferred truths. Lyu et al. \cite{Lyuetal21} learn embeddings for tasks, annotators, and labels by constructing heterogeneous networks from their interactions. These embeddings are iteratively refined based on the gradients of each sub-network. The TiReMGE model \cite{Wuetal2023a} takes task-annotator links as first-order interactions and treats task-task and annotator-annotator links as second-order interactions. The model estimates annotator reliability by minimizing the distance between encoded task representations and the labels provided by reliable annotators. The GOVERN model \cite{Liuetal24} is a graph contrastive learning framework built on an annotator-task bipartite graph, consisting of: a) a data augmentation module that selectively adds or drops edges; b) a bipartite graph neural network for node embeddings; and c) a contrastive objective combining both instance- and category-level losses to encourage robust, invariant embeddings. The CrowdFM model \cite{liu2026crowdfm} is a bipartite graph neural network that pretrains on a wide spectrum of simulated labeling scenarios to learn diverse annotation patterns and generalize across datasets without dataset-specific training.

Despite their powerful expressiveness, representation-learning-based methods require substantial data to train network parameters and degrade under data sparsity. Moreover, prior works build upon low-density annotator-task bipartite graphs, overlooking cross-annotator correlations that are more dense and encode intrinsic annotator reliability.
\section{Problem Definition}
We consider a crowdsourcing setting with $T$ tasks, $W$ annotators, and $K$ classes. Let $y_{ij}\in \{1\ldots K\}$ denote the label provided by annotator $w_{j}$ ($j\in\{1\ldots W\}$) for task $t_{i}$ ($i\in\{1\ldots T\}$), and $z_{i}\in \{1\ldots K\}$ denote the sole truth for $t_{i}$. Let $\mathcal{T}_{j}$ denote the set of tasks that $w_{j}$ has annotated, $\mathcal{W}_{i}$ denote the set of annotators who have annotated $t_{i}$, and $\mathcal{T}_{jq}=\mathcal{T}_{j}\cap\mathcal{T}_{q}$ denote the set of tasks annotated by both $w_{j}$ and $w_{q}$. The goal is to infer the latent truth $z_{i}$ for each $t_{i}$ solely from the noisy labels. This involves inferring: (i) the truth probability of each label, and (ii) the reliability of each annotator. Typically, most annotators label only a small subset of tasks.
\section{Proposed method: ahead}
In this section, we present the proposed \textbf{AHEAD} (cross-\textbf{A}nnotator learning and \textbf{H}igh-confid\textbf{E}nce \textbf{A}nnotator-guide\textbf{D} label aggregation), as illustrated in Figure \ref{framework}. Our objective is to accurately infer annotator reliability, especially bridging the estimation gap for annotators who label only a small subset of tasks from the population. To achieve this, we first extract annotator, truth distribution, and cross-annotator topology from the observed labels. Then, we model latent cross-annotator correlations to derive multi-view, complementary annotator embeddings, which are decoded into annotator-specific confusion matrices. We formulate a composite objective for effective model training. Finally, we update posterior truth probabilities.
\subsection{Feature Extraction}\label{feature-extract}
\textbf{Cross-Annotator Context Extraction.}
Prior works build upon the annotator-task bipartite graph, which is typically low-density as most individual annotators label only a small subset of tasks, while the number of tasks can be thousands of times greater than the number of annotators. A typical solution applies graph augmentation by selectively dropping or adding annotator-task edges; however, this may violate the intrinsic structure and requires high storage overhead, since edges often fail to satisfy the i.i.d. assumption \cite{Yangetal23}. In contrast, we focus on the intrinsic and dense cross-annotator contexts generated from the observed labels. This is because the cross-annotator topology graph, in which two annotators are connected whenever they share a labeled task, is typically much denser than the annotator-task bipartite graph induced by annotator individual labeling activities.

We extract the cross-annotator contextual topology $\mathcal{G}$, along with the positive and negative annotator pair sets, as follows:
\begin{myDef}\label{A-A-conn}
(Cross-Annotator Contextual Topology $\mathcal{G}$) For any annotator pair ($w_{j}$, $w_{q}$) satisfying $|\mathcal{T}_{j}\cap\mathcal{T}_{q}|\geq1$, we establish an undirected edge \{$w_{j}, w_{q}$\} between them and add it to $\mathcal{E}$, forming the cross-annotator contextual topology structure $\mathcal{G}=\{\{w_{j}\}_{j=1}^{W}, \mathcal{E}\}$. We define the contextual peers of $w_{j}$, denoted $\mathcal{N}(w_{j})$, as the nodes having an edge with $w_{j}$.
\end{myDef}
\begin{myDef}\label{pos-ng-pairs}
(Positive/Negative Annotator Pairs) For any annotator pair ($w_{j}$, $w_{q}$) satisfying $|\mathcal{T}_{jq}|=|\mathcal{T}_{j}\cap\mathcal{T}_{q}|\geq\delta_{0}$, we construct the positive and negative annotator pairs $\mathcal{E}_{+}$ and $\mathcal{E}_{-}$ according to Eq. (\ref{po-neg-pairs}),
\begin{equation}\label{po-neg-pairs}
    (w_{j},w_{q})\in
    \begin{cases}
        \mathcal{E}_{+}, & \text{if } \left(\frac{\sum_{t_{i}\in\mathcal{T}_{jq}}\mathbb{I}[y_{ij}=y_{iq}]}{|\mathcal{T}_{jq}|}-0.5\right)\cdot2 \geq \delta_{1} \\
        \mathcal{E}_{-}, & \text{if } \left(\frac{\sum_{t_{i}\in\mathcal{T}_{jq}}\mathbb{I}[y_{ij}=y_{iq}]}{|\mathcal{T}_{jq}|}-0.5\right)\cdot2 \leq -\delta_{2}
    \end{cases}
\end{equation}
where $\mathbb{I}[\cdot]$ is an indicator function that returns 1 if $[\cdot]$ is true and 0 otherwise. Moreover, we use $\mathcal{E}_{+}^{j}$ and $\mathcal{E}_{-}^{j}$ to denote the set of annotators that have a positive and a negative connection with $w_{j}$, respectively.
\end{myDef}

We extract $\mathcal{G}$, $\mathcal{E}_{+}$, and $\mathcal{E}_{-}$, which capture rich and intrinsic cross-annotator contexts that provide critical signals for annotator reliability estimation. Specifically, the cross-annotator topology $\mathcal{G}$ is much denser than annotator-task bipartite graph, enabling representation learning to model cross-annotator correlations and derive multi-view, complementary annotator embeddings by aggregating individual-level annotator features with contextual information. This dense topology also facilitates the propagation of reliability information across annotators. Meanwhile, we construct $\mathcal{E}_{+}$ and $\mathcal{E}_{-}$ to further refine annotator embeddings through contrastive learning.

\textbf{Teacher \& Initial Confusion Matrices.} Inspired by the Law of Large Numbers \cite{Durrett2010}, which suggests that empirical estimates converge to true values as sample size increases, we identify annotators whose answered task counts fall within a predefined top percentage as high-confidence annotators and treat their empirical confusion matrices as critical supervision. The motivation is to design a teacher-student loss that minimizes the discrepancy between the network-predicted confusion matrices of high-confidence annotators and their teacher confusion matrices, thereby guiding the network training.
\begin{myDef}\label{high-conf}
(High-Confidence Annotator Set $\mathcal{W}_{conf}$) We identify high-confidence annotators as those whose annotated task counts rank in the top $\rho$\%, and use $\mathcal{W}_{conf}$ to denote the high-confidence annotator set.
\end{myDef}
We derive teacher confusion matrices $\mathbf{C}^{(te)}$ of high-confidence annotators as follows. First, for each $t_{i}$ ($i\in\{1\ldots T\}$), we compute its pseudo-truth:
\begin{equation}
    \hat{z}_i = \arg\max_{k}\sum_{w_{j}\in\mathcal{W}_i} \mathbb{I}[y_{ij}=k]
\end{equation}
Then, for each $w_{j}\in\mathcal{W}_{conf}$, the corresponding teacher confusion matrix $\mathbf{C}^{(te)j}\in\mathbb{R}^{K \times K}$, with $\mathbf{C}^{(te)j}_{kl}$ denoting the probability that $w_{j}$ labels $l$ while the truth is $k$, is defined as:
\begin{equation}\label{teach-conf}
    \mathbf{C}^{(te)j}_{kl} =
    \begin{cases}
        \frac{\sum\limits_{t_{i}\in\mathcal{T}_{j}}\mathbb{I}[\hat{z}_i=k\land y_{ij}=l]}{\sum\limits_{t_{i}\in\mathcal{T}_{j}} \mathbb{I}[\hat{z}_i=k]}, & \text{if } \sum\limits_{t_{i}\in \mathcal{T}_{j}}\mathbb{I}[\hat{z}_i=k] > 0\\
        1, & \text{if } \sum\limits_{t_{i}\in \mathcal{T}_{j}}\mathbb{I}[\hat{z}_i=k] = 0 \land k=l\\
        0, & \text{otherwise}
    \end{cases}
\end{equation}

In contrast, the initial confusion matrix $\mathbf{C}^{(0)j}$ for each $w_{j}$ ($j=1\ldots W$), serving as the features to initialize annotator embeddings, is defined as:
\begin{equation}\label{remain-conf}
    \mathbf{C}^{(0)j}_{kl} =
    \begin{cases}
        \frac{\sum\limits_{t_{i}\in\mathcal{T}_{j}}\mathbb{I}[\hat{z}_i=k\land y_{ij}=l]}{\sum\limits_{t_{i}\in\mathcal{T}_{j}} \mathbb{I}[\hat{z}_i=k]}, & \text{if } \sum\limits_{t_{i}\in \mathcal{T}_{j}}\mathbb{I}[\hat{z}_i=k] > 0\\
        \delta_{3}, & \text{if } \sum\limits_{t_{i}\in\mathcal{T}_{j}}\mathbb{I}[\hat{z}_i=k] = 0 \land k=l\\
        0, & \text{otherwise}
    \end{cases}
\end{equation}
Notably, the confusion matrix for the same high-confidence annotator $w_{j}\in \mathcal{W}_{conf}$ is computed in two different ways depending on its role. For generating the initial annotator embedding ($\mathbf{C}^{(0)}$), we assign a conservative diagonal value of $\delta_{3}<1$ to all annotators for any class where they lack coverage (empty rows). In contrast, when used as the target supervisory signal ($\mathbf{C}^{(te)}$) in guiding the optimization trajectory, we assign a high diagonal value of 1.0 to high-confidence annotators. While this hard assignment implies a strong assumption, it is empirically justified in our context, as the selected high-confidence annotators typically exhibit comprehensive label coverage and rarely suffer from the missing class problem.

It should be noted that the supervisory guidance from high-confidence annotators differs substantially from prior methods relying on MV (Majority Voting)-derived pseudo-labels for cross-entropy optimization. Because while MV results are often noisy, reliability estimates for high-confidence annotators are statistically more robust, which is grounded in the Law of Large Numbers. This process not only reduces noise but also enables widespread propagation of supervision.

\textbf{Prior Truth Distribution.} Given the unsupervised setting and the Law of Large Numbers, we use the aggregated truth probabilities obtained from MV over all tasks to derive the prior truth distribution $\boldsymbol{\pi}$. Specifically, $\forall i \in \{1,\dots,T\}$, $k \in \{1,\dots,K\}$, we define
\begin{equation}
    \mathbf{\hat{z}}_{i,k} = \frac{\sum_{w_{j} \in \mathcal{W}_i} \mathbb{I}[y_{ij}=k]}{|\mathcal{W}_i|}
\end{equation}
\begin{equation}\label{prior-truth}
    \boldsymbol{\pi}_k = \frac{1}{T} \sum_{i=1}^{T} \mathbf{\hat{z}}_{i,k}
\end{equation}
\subsection{Cross-Annotator Learning}\label{A-A-GNN}
To derive robust and comprehensive annotator estimation in sparse crowdsourcing scenarios, we propose cross-annotator learning to derive multi-view, complementary annotator embeddings by aggregating individual-level annotator representations with their contextual information in $\mathcal{G}$. Then, we decode these embeddings into interpretable annotator-specific confusion matrices to fit the observations. This design reflects both the advantages of confusion-matrix-based and representation-learning-based modeling.

We derive initial annotator embeddings $\mathbf{h}^{(0)}\in\mathbb{R}^{W \times d_{1}}$ based on their initial confusion matrices $\mathbf{C}^{(0)}$, where $d_{1}$ denotes the hidden layer dimension. We first flat the initial confusion matrix of each annotator $\mathbf{C}^{(0)j}$ into a vector. These vectors are then projected into the embedding space using a learnable linear transformation matrix $\mathbf{A}\in \mathbb{R}^{K^2\times d_{1}}$ and a bias vector $\mathbf{b}\in \mathbb{R}^{d_{1}}$. Let $\mathcal{C}\in \mathbb{R}^{W \times K^2}$ denote the matrix reshaped from $\mathbf{C}^{(0)}$, where each row represents the flattened confusion matrix of an annotator. The initial annotator embedding $\mathbf{h}^{(0)}$ is derived as:
\begin{equation}
    \mathbf{h}^{(0)}=\mathcal{C}\mathbf{A}+\mathbf{b}
\end{equation}
$\forall w_{j}$ ($j\in\{1\ldots W\}$) and every $w_{q}\in\mathcal{N}(w_{j})$, we learn an attention coefficient between the source node $w_{q}$ and the target node $w_{j}$. Specifically, we first concatenate the projected features of $w_{q}$ and $w_{j}$, and then encode the combined representation to the hidden dimension $d_{1}$ using a learnable weight matrix $\mathbf{A}_{gat}\in\mathbb{R}^{d_{1} \times 2d_{1}}$, followed by a LeakyReLU activation. This representation is then projected onto a learnable weight vector $\mathbf{a}_{gat}\in\mathbb{R}^{d_{1}}$ to yield the attention score $x_{qj}$. Namely,
\begin{equation}
    x_{qj} = \mathbf{a}_{gat}^{\mathsf{T}}\text{LeakyReLU}\left( \mathbf{A}_{gat} \left[\mathbf{h}_q^{(s-1)} \, \| \, \mathbf{h}_j^{(s-1)}\right] \right)
\end{equation}
where $\mathbf{h}_q^{(s-1)}$ denotes the embedding of $w_{q}$ at iteration $s-1$ and $\|$ denotes the concatenation operation. The attention scores are then normalized to derive the attention coefficient $\alpha_{qj}$:
\begin{equation}
    \alpha_{qj} = \frac{e^{x_{qj}}}{\sum_{w_{q'} \in \mathcal{N}(w_{j})} e^{x_{q'j}}}
\end{equation}
Next, we update $w_{j}$'s representation $\mathbf{h}_j^{(s)}$ in iteration $s$ by aggregating its own features with contextual messages from its topologically connected annotators, using a scaling factor $\lambda_{0}$ to stabilize training:
\begin{equation}\label{hsj}
    \mathbf{h}_{j}^{(s)} = \mathbf{h}_{j}^{(s-1)}+\lambda_{0}\sum_{w_{q'}\in \mathcal{N}(w_{j})} \alpha_{q'j}\mathbf{h}_{q'}^{(s-1)}
\end{equation}
The updated annotator representation $\mathbf{h}_j^{(s)}$ is then decoded to predict annotator-specific confusion matrix $\mathbf{C}^{(s)j}$. The decoder is implemented as a Multi-Layer Perceptron (MLP) parameterized by two linear layers ($\mathbf{A}_{1}\in \mathbb{R}^{d_{1}\times d_{1}}$, $\mathbf{b}_{1}\in \mathbb{R}^{d_{1}}$), ($\mathbf{A}_{2}\in \mathbb{R}^{K^{2}\times d_{1}}$, $\mathbf{b}_{2}\in \mathbb{R}^{K^{2}}$), and a ReLU activation, projecting the latent embedding $\mathbf{h}_j^{(s)}$ to the output $\mathbf{u}_j \in \mathbb{R}^{K^2}$ as follows:
\begin{equation}
    \mathbf{u}_{j} = \mathbf{A}_{2} \left( \text{ReLU}(\mathbf{A}_{1} \mathbf{h}_{j}^{(s)} + \mathbf{b}_{1}) \right) + \mathbf{b}_{2}
\end{equation}
We reshape $\mathbf{u}_j$ into a matrix $\mathbf{M}^j \in \mathbb{R}^{K \times K}$, and apply a row-wise Softmax to obtain the predicted confusion matrix $\mathbf{C}^{(s)j}$:
\begin{equation}\label{csj}
\mathbf{C}^{(s)j}_{kl}=\frac{e^{\mathbf{M}^j_{kl}}}{\sum_{l'=1}^{K} e^{\mathbf{M}^j_{kl'}}}
\end{equation}
where $\mathbf{C}^{(s)j}_{kl}$ denotes the probability that $w_{j}$ labels $l$ while the truth is $k$. Our correlation-aware annotator embeddings enable annotators with sparse labels to borrow statistical strength from the population, thereby stabilizing reliability estimation in scenarios where independent models fail. Crucially, the predicted annotator confusion matrices $\mathbf{C}^{(s)}$ provide a principled formulation for the label generation process, thereby effectively integrating the advantages of both confusion-matrix-based and representation-learning-based paradigms. For efficient model training, we formulate a composite objective that integrates likelihood maximization with structural regularization and directional guidance. The optimization process operates within an Expectation-Maximization (EM) framework: refining posterior truth probabilities (E-step) and updating the model parameters (M-step). The total loss function $\mathcal{L}_{total}$ comprises four components: (1) a negative log-likelihood term to capture observation consistency; (2) a teacher-student loss from high-confidence annotators to guide the optimization trajectory; (3) a graph-based contrastive loss to refine the training of annotator embeddings; (4) a regularization term to prevent degenerate solutions.
\subsection{Negative Log-Likelihood Loss}
We introduce the negative log-likelihood loss $\mathcal{L}_{Nll}$ to optimize the network parameters by minimizing the expected negative log-likelihood regarding the posterior truth probabilities. Let $\mathbf{\hat{z}}_{ik}=P(z_i=k|Y,\mathbf{C}^{(s)})$ denote the probability that class $k$ being the truth of $t_{i}$, $\mathcal{L}_{Nll}$ is defined as:
\begin{equation}\label{loss-nll}
    \mathcal{L}_{Nll}=-\frac{1}{T} \sum_{i=1}^{T}\sum_{w_{j}\in\mathcal{W}_{i}}\sum_{k=1}^{K} \mathbf{\hat{z}}_{ik}\cdot\log \mathbf{C}^{(s)j}_{ky_{ij}}
\end{equation}
By parameterizing confusion matrices through a graph neural network rather than estimating them as independent variables, we introduce a shared representation that allows the model to generalize across annotators. The rationale of $\mathcal{L}_{Nll}$ is to optimize network parameters by maximizing the plausibility of observed crowdsourced labels conditioned on the updated posterior truth probabilities. By quantifying the discrepancy between the predicted annotator confusion matrices and the actual annotator labels, $\mathcal{L}_{Nll}$ effectively enforces a consistency constraint that aligns the predicted confusion matrices with the statistical consensus of the crowd. Therefore, the network iteratively refines its parameters to fit the current posterior truth probabilities, thereby allowing the system to proceed toward an accurate aggregation of labels without the prerequisite of external supervision.
\subsection{Teacher-Student Loss for High-Confidence Annotators}
Relying solely on internal consensus can cause the model to oscillate or converge to local optima. Therefore, the teacher-student Loss $\mathcal{L}_{Teacher}$ is designed to guide the optimization trajectory by introducing reliable structural supervision from high-confidence annotators. By treating the empirical confusion matrices of high-confidence annotators as highly reliable, we guide the model by minimizing the discrepancy between their predicted confusion matrices and the empirical counterparts. For each annotator $w_{j}\in\mathcal{W}_{conf}$, given the predicted confusion matrix $\mathbf{C}^{(s)j}$, $\mathcal{L}_{Teacher}$ is defined as:
\begin{equation}\label{loss-teacher}
    \mathcal{L}_{Teacher} = \frac{1}{|\mathcal{W}_{conf}|} \sum_{w_{j}\in\mathcal{W}_{conf}}\left\|\mathbf{C}^{(te)j}-\mathbf{C}^{(s)j}\right\|_{F}^{2}
\end{equation}
This objective offers two distinct benefits. First, by minimizing the mean squared error between the predicted and empirical confusion matrices of high-confidence annotators, $\mathcal{L}_{Teacher}$ injects strong supervision into the network, effectively mitigating the identifiability issues inherent in unsupervised EM. Second, the extensive task coverage of high-confidence annotators endows them with dense connectivity with other annotators within the topology, which facilitates annotator reliability propagation and benefit annotators with sparse data.
\subsection{Graph-Based Contrastive Loss}
To enforce discriminative representation learning, we introduce the graph-based contrastive loss $\mathcal{L}_{Ctr}$, which explicitly translates the structural positive and negative annotator pairs into strict geometric constraints within the embedding space. Formally, $\mathcal{L}_{Ctr}$ is defined as follows:
\begin{equation}\label{loss-contrast}
    \mathcal{L}_{Ctr} = \frac{-1}{|\mathcal{E}_{+}|}\sum_{(w_{j},w_{q})\in\mathcal{E}_{+}}\log\frac{e^{\frac{r(\mathbf{h}_{j}^{(s)}, \mathbf{h}_{q}^{(s)})}{\tau}}}{e^{\frac{r(\mathbf{h}_{j}^{(s)}, \mathbf{h}_{q}^{(s)})}{\tau}}+\sum\limits_{w_{q'}\in\mathcal{E}_{-}^{j}}e^{\frac{r(\mathbf{h}_{j}^{(s)}, \mathbf{h}_{q'}^{(s)})}{\tau}}}
\end{equation}
where $\tau$ is a temperature parameter and $r$ denotes the similarity score function that measures the similarity between node embeddings, defined as:
\begin{equation}\label{loss-contrast-similarity}
     r(\mathbf{h}_{j}^{(s)},\mathbf{h}_{q}^{(s)})=\frac{(\mathbf{h}_{j}^{(s)})\cdot(\mathbf{h}_{q}^{(s)})^{\mathsf{T}}}{\|\mathbf{h}_{j}^{(s)}\|\cdot\|\mathbf{h}_{q}^{(s)}\|}
\end{equation}
By pulling positive annotator pairs together while repelling negative ones apart, $\mathcal{L}_{Ctr}$ prevents feature collapse and ensures a clear separation between reliable and unreliable annotators. The efficacy of this formulation lies in hard negative mining. The temperature-scaled softmax denominator assigns significantly larger gradients to the most difficult negative pairs, directing the optimization focus toward resolving difficult ambiguities rather than trivial samples. Furthermore, this module serves as a structural regularizer that maximizes the mutual information between the representations of positive annotator pairs, yielding robust, separable embeddings that generalize effectively to unseen data.
\subsection{Regularization Loss}
Finally, to prevent convergence to degenerate solutions, we introduce a weak prior on annotator reliability by defining a regularization term $\mathcal{L}_{Reg}$ that encourages confusion matrices to be close to the identity matrix $\mathbf{I}$:
\begin{equation}\label{loss-reg}
    \mathcal{L}_{Reg} = \frac{1}{W} \sum_{j=1}^{W}\left\|\mathbf{C}^{(s)j} - \mathbf{I} \right\|_{F}^{2}
\end{equation}
The primary role of $\mathcal{L}_{Reg}$ is to impose a diagonal dominance constraint on confusion matrices. By grounding the optimization in the prior assumption that annotators are generally reliable, this regularization ensures that the learned confusion matrices deviate from the diagonal only when data-driven evidence is strong enough to override the penalty, thereby capturing genuine annotator bias rather than drifting due to the inherent instability of unsupervised learning.
\subsection{Truth Estimation}
With the above four loss terms established, each tailored to capture a specific aspect of the learning process and provide complementary supervision signals, the total objective function $\mathcal{L}_{total}$ is a weighted combination of these components:
\begin{equation}\label{loss-total}
    \mathcal{L}_{total} = \lambda_{1}\mathcal{L}_{Nll} + \lambda_{2} \mathcal{L}_{Ctr} + \lambda_{3} \mathcal{L}_{Teacher} + \lambda_{4} \mathcal{L}_{Reg}
\end{equation}
where $\lambda_{1}$, $\lambda_{2}$, $\lambda_{3}$, and $\lambda_{4}$ are loss weights that control the relative importance of different supervision signals. $\mathcal{L}_{total}$ balances data likelihood with structural constraints and prior knowledge. We optimize this objective using the Adam optimizer.

Based on the predicted annotator-specific confusion matrices, we update the posterior truth probabilities as follows:
\begin{equation}\label{update-Z}
    \mathbf{\hat{z}}_{ik}\propto\exp\left(\sum_{w_{j}\in\mathcal{W}_i}\log (\mathbf{C}_{ky_{ij}}^{(s)j})+\lambda_{5}\log\boldsymbol{\pi}_k\right)
\end{equation}
where $\lambda_{5}$ is a scaling factor to the prior term. Finally, we treat the label with the highest probability as the truth.
\begin{algorithm}[t]
\LinesNumbered
\caption{AHEAD Training Algorithm}
\label{alg:ahead_training}
\SetKwInOut{Input}{Input}
\SetKwInOut{Output}{Output}

\Input{
    Label set $Y$, epochs $S$, learning rate $\eta$, hidden layer dimension $d_{1}$, temperature $\tau$, loss weights $\lambda_{1}, \lambda_{2}, \lambda_{3}, \lambda_{4}$, thresholds $\rho, \delta_{0}, \delta_{1}, \delta_{2}, \delta_{3}$, and scaling factors $\lambda_{0}$, $\lambda_{5}$.
}
Generate cross-annotator topology $\mathcal{G} = \{\{w_{j}\}_{j=1}^{W}, \mathcal{E}\}$;\\
Generate pos/neg annotator pairs $\mathcal{E}_{+}$, $\mathcal{E}_{-}$ using Eq. (\ref{po-neg-pairs});\\
Generate teacher/initial confusion matrices $\mathbf{C}^{(te)}$, $\mathbf{C}^{(0)}$ using Eq. (\ref{teach-conf}) and Eq. (\ref{remain-conf});\\
Generate prior truth distribution $\boldsymbol{\pi}$ using Eq. (\ref{prior-truth});\\
\For{epoch $s = 1$ to $S$}{
    Perform forward pass: $\mathbf{h}^{(s)}, \mathbf{C}^{(s)} = \text{GNN}(\mathcal{G}, \mathbf{C}^{(0)}; \Theta)$ using Eq. (\ref{hsj}) and Eq. (\ref{csj}), respectively;\\

    Calculate $\mathcal{L}_{Nll}$ using Eq. (\ref{loss-nll});\\
    Calculate $\mathcal{L}_{Teacher}$ using Eq. (\ref{loss-teacher});\\
    Calculate $\mathcal{L}_{Ctr}$ using Eq. (\ref{loss-contrast});\\
    Calculate $\mathcal{L}_{Reg}$ using Eq. (\ref{loss-reg});\\
    Calculate the final loss $\mathcal{L}_{total}$ using Eq. (\ref{loss-total});\\
    Update $\Theta$ by minimizing $\mathcal{L}_{total}$ using Adam optimizer;\\
    Update truth probabilities $\hat{Z}$ using Eq. (\ref{update-Z});\\
}
\Output{
    $\hat{Z}$.
}
\end{algorithm}
\begin{table*}[htbp]
\renewcommand{\arraystretch}{1.1}
\centering
\caption{Dataset statistics. $Min_{w}$ and $Med_{w}$ denote the minimum and median numbers of tasks labeled across annotators.}
\label{dataset}
\begin{tabularx}{\textwidth}{lllYYYYlY}
\noalign{\hrule height 1pt}
Dataset & $T$ & $W$ & $K$ & $Min_{w}$ & $Med_{w}$ & \#Labels & Truth distribution & Domain \\
\noalign{\hrule height 0.7pt}
Val7    & 100   & 38  & 7 & 20 & 20  & 1{,}000  & \{0.08,0.21,0.18,0.10,0.23,0.19,0.01\} & Sentiment \\
Aircr   & 593   & 50  & 6 & 1 & 4.5  & 1{,}588  & \{0.17,0.17,0.13,0.17,0.17,0.19\} & Image \\
CF      & 300   & 461 & 5 & 1 & 2  & 1{,}720  & \{0.19,0.23,0.24,0.31,0.03\} & Sentiment \\
MS      & 700   & 44  & 10 & 2 & 16  & 2{,}945  & \{0.11,0.10,0.10,0.09,0.10,0.10,0.10,0.11,0.11,0.09\} & Audio \\
Dog     & 807   & 109  & 4 & 1 & 38  & 8{,}070  & \{0.21,0.23,0.27,0.29\} & Image \\
Face    & 584   & 27  & 4 & 4 & 56  & 5{,}242  & \{0.25,0.25,0.25,0.25\} & Image \\
Adult   & 11{,}040 & 825 & 4 & 1 & 18 & 89{,}799  & \{0.56,0.18,0.11,0.15\} & Level \\
Val5    & 100   & 38  & 5 & 20 & 20  & 1{,}000  & \{0.13,0.27,0.23,0.28,0.09\} & Sentiment \\
Web     & 2{,}665 & 177  & 5 & 1 & 19  & 15{,}567  & \{0.12,0.18,0.26,0.20,0.23\} & Relevance \\
LabelMe & 1{,}000 & 59  & 8 & 3 & 27 & 2{,}547  & \{0.09,0.12,0.13,0.11,0.14,0.13,0.13,0.15\} & Image \\
\noalign{\hrule height 0.7pt}
Senti & 98{,}980 & 1{,}960  & 5 & 8 & 45 & 569{,}274  & \{0.23,0.25,0.19,0.31,0.02\} & Sentiment \\
\noalign{\hrule height 1pt}
\end{tabularx}
\end{table*}
\subsection{Algorithm}
Let $\Theta$ include all trainable parameters in our graph neural network. Algorithm \ref{alg:ahead_training} outlines the procedure of AHEAD. The algorithm operates in an iterative Expectation-Maximization (EM) manner. In the initialization phase, we construct an cross-annotator topology $\mathcal{G}$ from their interactions, generate teacher confusion matrices for high-confidence annotators, calculate initial confusion matrices for all annotators, and derive prior truth distribution. During each training epoch, we run a graph neural network to learn cross-annotator attention coefficients over their topology structure, deriving comprehensive annotator embeddings by aggregating each annotator's own representation with contextual messages from its connected annotators. These embeddings are subsequently decoded into interpretable annotator-specific confusion matrices to fit the observed labels. The model parameters $\Theta$ are updated by minimizing a composite loss $\mathcal{L}_{total}$, which consists of the negative log-likelihood, a teacher-student loss for high-confidence annotators, a graph-based contrastive learning loss, and a regularization term. Then, the posterior truth probabilities $\hat{Z}$ are updated based on the predicted confusion matrices.

AHEAD combines both the strengths of representation learning and confusion-matrix-based modeling. By modeling dense cross-annotator learning over their interaction topology, the model effectively captures rich, intrinsic cross-annotator correlations and propagates annotator reliability across the network. This correlation-aware annotator modeling enables annotators with sparse labels to leverage population-level information. Meanwhile, the decoded confusion matrices provide a theoretically interpretable fit of the label generation, ensuring interpretable cross-annotator learning and effective model training.
\section{Experiments}
\subsection{Experimental Setup}
\noindent
\textbf{Datasets.} We use 11 widely adopted real-world datasets spanning five domains for evaluation. Table \ref{dataset} summarizes their statistics. Among them, CF, MS, Dog, Face, Adult, Web, and Senti are also used in \cite{Lietal2019b}; Val5, Val7, and Aircr are also used in \cite{Wuetal2023b}; and LabelMe also appears in \cite{Liuetal24}. Notably, the first 10 datasets are used to evaluate overall performance, whereas the last dataset, Senti, being the largest publicly available dataset, is used to evaluate scalability.

\noindent
\textbf{Metrics.} We evaluate the results of \textit{Accuracy} and \textit{Macro-F1}.

\noindent
\textbf{Implementation details.} We set $\delta_{0}=4$, $\delta_{1}=0.5$, $\delta_{2}=0.4$, $\delta_{3}=0.2$, $\rho=30$, $S=100$, $\eta=0.01$, $d_{1}=128$, $\tau=0.2$, $\lambda_{0}=0.6$, $\lambda_{1}=1$, $\lambda_{2}=1$, $\lambda_{3}=6$, and $\lambda_{4}=0.5$. For $\lambda_{5}$, we set $\lambda_{5}=1$ during the first 20\% of the epochs, and $\lambda_{5}=0.00001T$ for the remaining epochs.

\noindent
\textbf{Baselines.} We compare AHEAD with four \textit{single-parameter-based} (MV, CATD \cite{Lietal2014}, GLAD \cite{Whitehilletal2009}, BWA \cite{Lietal2019b}), four \textit{confusion-matrix-based} (DS \cite{DawidSkene1979}, IBCC \cite{KimGhahramani2012}, CBCC \cite{Venanzietal2014}, EBCC \cite{Lietal2019a}), and three \textit{representation-learning-based} (TiReMGE \cite{Wuetal2023a}, GOVERN \cite{Liuetal24}, CrowdFM \cite{liu2026crowdfm}) methods. Their detailed descriptions can be found in Section \ref{related-work}.

All experiments were conducted on a server with 2 vCPUs and 8 GB of memory, and repeated 10 times.
\begin{table*}[t]
\renewcommand{\arraystretch}{1.25}
\centering
\setlength{\tabcolsep}{2.1pt}
\caption{Performance results across different datasets and methods.}
\label{tab:performance}
\begin{tabularx}{\textwidth}{lYYYYYYYYYY>{\columncolor{gray!40}}Y}
\noalign{\hrule height 1pt}
\noalign{\vskip 4pt}
\multicolumn{12}{c}{\textit{Accuracy (mean\% \(\pm\) std.\%)} \(\uparrow\)} \\
\noalign{\hrule height 0.8pt}
Method & Val7 & Aircr & CF & MS & Dog & Face & Adult & Val5 & Web & LabelMe & Avg. \\
\noalign{\hrule height 0.8pt}
MV      & 19.90{\footnotesize$\pm$0.94} & 81.01{\footnotesize$\pm$0.65} & 88.73{\footnotesize$\pm$0.53} & 70.17{\footnotesize$\pm$0.65} & 82.33{\footnotesize$\pm$0.29} & 63.61{\footnotesize$\pm$0.49} & 75.86{\footnotesize$\pm$0.49} & 35.30{\footnotesize$\pm$2.83} & 73.19{\footnotesize$\pm$0.57} & 76.55{\footnotesize$\pm$0.40} & 66.67{\footnotesize$\pm$0.33} \\
CATD    & 16.00{\footnotesize$\pm$0.00} & 80.29{\footnotesize$\pm$0.46} & 87.47{\footnotesize$\pm$0.86} & 78.69{\footnotesize$\pm$0.59} & 82.11{\footnotesize$\pm$0.22} & 61.66{\footnotesize$\pm$0.05} & 75.08{\footnotesize$\pm$0.23} & 33.00{\footnotesize$\pm$0.00} & 80.32{\footnotesize$\pm$0.05} & 76.78{\footnotesize$\pm$0.36} & 67.14{\footnotesize$\pm$0.17} \\
GLAD    & 20.00{\footnotesize$\pm$0.00} & 80.61{\footnotesize$\pm$0.00} & 88.00{\footnotesize$\pm$0.00} & 78.43{\footnotesize$\pm$0.00} & 83.52{\footnotesize$\pm$0.00} & 63.01{\footnotesize$\pm$0.00} & 76.28{\footnotesize$\pm$0.00} & 33.00{\footnotesize$\pm$0.00} & 79.87{\footnotesize$\pm$0.00} & 77.00{\footnotesize$\pm$0.00} & 67.97{\footnotesize$\pm$0.00} \\
BWA     & 20.00{\footnotesize$\pm$0.00} & 82.46{\footnotesize$\pm$0.00} & 89.33{\footnotesize$\pm$0.00} & 78.57{\footnotesize$\pm$0.00} & 83.15{\footnotesize$\pm$0.00} & 61.82{\footnotesize$\pm$0.00} & 74.17{\footnotesize$\pm$0.00} & 35.00{\footnotesize$\pm$0.00} & 82.25{\footnotesize$\pm$0.00} & 77.40{\footnotesize$\pm$0.00} & 68.42{\footnotesize$\pm$0.00} \\
\midrule
DS      & \underline{22.00{\footnotesize$\pm$0.00}} & \underline{84.15{\footnotesize$\pm$0.00}} & 79.67{\footnotesize$\pm$0.00} & 76.43{\footnotesize$\pm$0.00} & \textbf{84.26{\footnotesize$\pm$0.00}} & 64.04{\footnotesize$\pm$0.00} & 73.87{\footnotesize$\pm$0.00} & \underline{41.00{\footnotesize$\pm$0.00}} & 82.55{\footnotesize$\pm$0.00} & \underline{79.30{\footnotesize$\pm$0.00}} & 68.73{\footnotesize$\pm$0.00} \\
IBCC    & 17.00{\footnotesize$\pm$0.00} & 78.08{\footnotesize$\pm$0.00} & 88.33{\footnotesize$\pm$0.00} & 76.43{\footnotesize$\pm$0.00} & 83.77{\footnotesize$\pm$0.00} & 64.73{\footnotesize$\pm$0.00} & \underline{77.18{\footnotesize$\pm$0.00}} & 33.00{\footnotesize$\pm$0.00} & 70.41{\footnotesize$\pm$0.00} & 77.50{\footnotesize$\pm$0.00} & 66.64{\footnotesize$\pm$0.00} \\
CBCC    & 13.00{\footnotesize$\pm$0.00} & 76.56{\footnotesize$\pm$0.00} & 89.33{\footnotesize$\pm$0.00} & 77.29{\footnotesize$\pm$0.00} & 80.79{\footnotesize$\pm$0.00} & 65.41{\footnotesize$\pm$0.00} & 75.08{\footnotesize$\pm$0.00} & 33.00{\footnotesize$\pm$0.00} & 70.67{\footnotesize$\pm$0.00} & 76.50{\footnotesize$\pm$0.00} & 65.76{\footnotesize$\pm$0.00} \\
EBCC    & 16.70{\footnotesize$\pm$0.78} & 81.32{\footnotesize$\pm$0.07} & 88.33{\footnotesize$\pm$0.00} & 78.71{\footnotesize$\pm$0.00} & 84.01{\footnotesize$\pm$0.00} & 63.46{\footnotesize$\pm$0.78} & 74.77{\footnotesize$\pm$0.13} & 33.40{\footnotesize$\pm$0.49} & 74.37{\footnotesize$\pm$0.00} & 78.60{\footnotesize$\pm$0.00} & 67.37{\footnotesize$\pm$0.15} \\
\midrule
TiReMGE & \underline{22.00{\footnotesize$\pm$0.45}} & 80.25{\footnotesize$\pm$0.19} & 88.90{\footnotesize$\pm$0.26} & 72.09{\footnotesize$\pm$0.16} & 82.49{\footnotesize$\pm$0.32} & 64.32{\footnotesize$\pm$0.24} & 76.58{\footnotesize$\pm$0.00} & 37.40{\footnotesize$\pm$0.49} & 67.03{\footnotesize$\pm$0.71} & 77.15{\footnotesize$\pm$0.17} & 66.82{\footnotesize$\pm$0.09} \\
GOVERN  & 14.70{\footnotesize$\pm$2.19} & 81.85{\footnotesize$\pm$0.46} & \underline{89.67{\footnotesize$\pm$0.84}} & 77.83{\footnotesize$\pm$0.50} & 82.29{\footnotesize$\pm$0.90} & \textbf{66.56{\footnotesize$\pm$0.88}} & 69.52{\footnotesize$\pm$1.49} & 32.10{\footnotesize$\pm$2.77} & \textbf{89.66{\footnotesize$\pm$1.70}} & 77.56{\footnotesize$\pm$0.83} & 68.17{\footnotesize$\pm$0.53} \\
CrowdFM  & 20.40{\footnotesize$\pm$1.96} & 81.82{\footnotesize$\pm$0.56} & 88.00{\footnotesize$\pm$0.77} & \underline{79.61{\footnotesize$\pm$0.36}} & 82.65{\footnotesize$\pm$0.57} & 63.82{\footnotesize$\pm$0.67} & 75.95{\footnotesize$\pm$0.51} & 33.40{\footnotesize$\pm$2.46} & \underline{84.71{\footnotesize$\pm$3.00}} & 77.12{\footnotesize$\pm$0.51} & \underline{68.75{\footnotesize$\pm$0.73}} \\
AHEAD   & \textbf{36.90{\footnotesize$\pm$2.21}} & \textbf{85.06{\footnotesize$\pm$0.31}} & \textbf{89.77{\footnotesize$\pm$0.30}} & \textbf{79.79{\footnotesize$\pm$1.01}} & \underline{84.15{\footnotesize$\pm$0.21}} & \underline{65.84{\footnotesize$\pm$0.16}} & \textbf{77.54{\footnotesize$\pm$0.40}} & \textbf{49.40{\footnotesize$\pm$2.20}} & 83.81{\footnotesize$\pm$0.36} & \textbf{80.03{\footnotesize$\pm$0.37}} & \textbf{73.23{\footnotesize$\pm$0.37}} \\
\noalign{\hrule height 1pt}
\noalign{\vskip 4pt}
\multicolumn{12}{c}{\textit{Macro-F1 (mean\% \(\pm\) std.\%)} \(\uparrow\)} \\
\noalign{\hrule height 0.8pt}
Method & Val7 & Aircr & CF & MS & Dog & Face & Adult & Val5 & Web & LabelMe & Avg. \\
\noalign{\hrule height 0.8pt}
MV      & 17.80{\footnotesize$\pm$0.83} & 79.73{\footnotesize$\pm$0.71} & 79.11{\footnotesize$\pm$1.60} & 70.37{\footnotesize$\pm$0.30} & 82.13{\footnotesize$\pm$0.37} & 61.88{\footnotesize$\pm$0.60} & 63.02{\footnotesize$\pm$0.83} & 31.61{\footnotesize$\pm$1.90} & 72.93{\footnotesize$\pm$0.40} & 76.09{\footnotesize$\pm$0.54} & 63.47{\footnotesize$\pm$0.29} \\
CATD    & 9.44{\footnotesize$\pm$0.00} & 79.06{\footnotesize$\pm$0.48} & 77.98{\footnotesize$\pm$1.37} & 79.21{\footnotesize$\pm$0.46} & 82.13{\footnotesize$\pm$0.23} & 59.36{\footnotesize$\pm$0.07} & 62.50{\footnotesize$\pm$0.56} & 26.10{\footnotesize$\pm$0.00} & 80.41{\footnotesize$\pm$0.05} & 76.29{\footnotesize$\pm$0.36} & 63.25{\footnotesize$\pm$0.23} \\
GLAD    & 16.90{\footnotesize$\pm$0.00} & 79.47{\footnotesize$\pm$0.00} & 77.98{\footnotesize$\pm$0.00} & 78.81{\footnotesize$\pm$0.00} & 83.45{\footnotesize$\pm$0.00} & 61.04{\footnotesize$\pm$0.00} & 63.43{\footnotesize$\pm$0.00} & 29.57{\footnotesize$\pm$0.00} & 80.30{\footnotesize$\pm$0.00} & 76.57{\footnotesize$\pm$0.00} & 64.75{\footnotesize$\pm$0.00} \\
BWA     & 15.80{\footnotesize$\pm$0.00} & 81.45{\footnotesize$\pm$0.00} & \underline{79.36{\footnotesize$\pm$0.00}} & 78.98{\footnotesize$\pm$0.00} & 83.15{\footnotesize$\pm$0.00} & 59.47{\footnotesize$\pm$0.00} & 59.88{\footnotesize$\pm$0.00} & 28.51{\footnotesize$\pm$0.00} & 82.18{\footnotesize$\pm$0.00} & 76.84{\footnotesize$\pm$0.00} & 64.56{\footnotesize$\pm$0.00} \\
\midrule
DS      & \underline{21.00{\footnotesize$\pm$0.00}} & \underline{83.06{\footnotesize$\pm$0.00}} & 71.51{\footnotesize$\pm$0.00} & 76.61{\footnotesize$\pm$0.00} & \textbf{84.46{\footnotesize$\pm$0.00}} & 62.63{\footnotesize$\pm$0.00} & 62.02{\footnotesize$\pm$0.00} & \underline{41.75{\footnotesize$\pm$0.00}} & 82.35{\footnotesize$\pm$0.00} & \underline{78.67{\footnotesize$\pm$0.00}} & \underline{66.41{\footnotesize$\pm$0.00}} \\
IBCC    & 10.47{\footnotesize$\pm$0.00} & 73.92{\footnotesize$\pm$0.00} & 71.36{\footnotesize$\pm$0.00} & 76.59{\footnotesize$\pm$0.00} & 83.87{\footnotesize$\pm$0.00} & 63.34{\footnotesize$\pm$0.00} & \underline{65.53{\footnotesize$\pm$0.00}} & 21.65{\footnotesize$\pm$0.00} & 69.77{\footnotesize$\pm$0.00} & 77.07{\footnotesize$\pm$0.00} & 61.36{\footnotesize$\pm$0.00} \\
CBCC    & 8.87{\footnotesize$\pm$0.00} & 70.81{\footnotesize$\pm$0.00} & 75.99{\footnotesize$\pm$0.00} & 77.04{\footnotesize$\pm$0.00} & 80.54{\footnotesize$\pm$0.00} & 64.35{\footnotesize$\pm$0.00} & 64.26{\footnotesize$\pm$0.00} & 21.65{\footnotesize$\pm$0.00} & 68.00{\footnotesize$\pm$0.00} & 75.43{\footnotesize$\pm$0.00} & 60.69{\footnotesize$\pm$0.00} \\
EBCC    & 10.29{\footnotesize$\pm$0.48} & 79.48{\footnotesize$\pm$0.06} & 71.44{\footnotesize$\pm$0.00} & 79.05{\footnotesize$\pm$0.00} & 84.17{\footnotesize$\pm$0.00} & 61.26{\footnotesize$\pm$1.05} & 62.08{\footnotesize$\pm$0.37} & 22.89{\footnotesize$\pm$2.02} & 74.40{\footnotesize$\pm$0.00} & 78.18{\footnotesize$\pm$0.00} & 62.32{\footnotesize$\pm$0.24} \\
\midrule
TiReMGE & 18.77{\footnotesize$\pm$4.50} & 78.82{\footnotesize$\pm$0.28} & 76.20{\footnotesize$\pm$1.21} & 72.22{\footnotesize$\pm$0.27} & 82.47{\footnotesize$\pm$0.35} & 62.43{\footnotesize$\pm$0.38} & 64.32{\footnotesize$\pm$0.07} & 33.99{\footnotesize$\pm$1.24} & 67.54{\footnotesize$\pm$0.68} & 76.85{\footnotesize$\pm$0.25} & 63.36{\footnotesize$\pm$0.54} \\
GOVERN  & 9.40{\footnotesize$\pm$2.12} & 80.36{\footnotesize$\pm$0.57} & 78.02{\footnotesize$\pm$3.57} & 78.22{\footnotesize$\pm$0.48} & 82.18{\footnotesize$\pm$1.03} & \textbf{65.48{\footnotesize$\pm$1.09}} & 43.28{\footnotesize$\pm$4.02} & 25.38{\footnotesize$\pm$4.75} & \textbf{89.61{\footnotesize$\pm$1.79}} & 76.89{\footnotesize$\pm$0.77} & 62.88{\footnotesize$\pm$0.80} \\
CrowdFM  & 17.37{\footnotesize$\pm$2.63} & 80.68{\footnotesize$\pm$0.57} & 77.40{\footnotesize$\pm$1.41} & \underline{80.01{\footnotesize$\pm$0.35}} & 82.60{\footnotesize$\pm$0.61} & 61.88{\footnotesize$\pm$0.84} & 63.31{\footnotesize$\pm$0.90} & 28.08{\footnotesize$\pm$3.68} & \underline{84.81{\footnotesize$\pm$3.05}} & 76.61{\footnotesize$\pm$0.55} & 65.28{\footnotesize$\pm$0.88} \\
AHEAD   & \textbf{33.25{\footnotesize$\pm$2.00}} & \textbf{84.01{\footnotesize$\pm$0.36}} & \textbf{81.11{\footnotesize$\pm$0.98}} & \textbf{80.02{\footnotesize$\pm$1.03}} & \underline{84.38{\footnotesize$\pm$0.20}} & \underline{65.03{\footnotesize$\pm$0.18}} & \textbf{68.99{\footnotesize$\pm$0.50}} & \textbf{48.06{\footnotesize$\pm$2.43}} & 83.31{\footnotesize$\pm$0.37} & \textbf{79.67{\footnotesize$\pm$0.39}} & \textbf{70.78{\footnotesize$\pm$0.39}} \\
\noalign{\hrule height 1pt}
\end{tabularx}
\end{table*}
\subsection{Effectiveness}
Table \ref{tab:performance} presents the \textit{Accuracy} and \textit{Macro-F1} results of 12 methods across 10 real-world datasets. Figure \ref{fig-radar} compares the class-wise accuracy of AHEAD with representative confusion-matrix-based methods (DS and IBCC) and representation-learning-based methods (TiReMGE, GOVERN, and CrowdFM).

Table \ref{tab:performance} demonstrates that AHEAD consistently secures top-tier performance across all datasets, improving the average accuracy and Macro-F1 score from 68.75\% to 73.23\%, and from 66.41\% to 70.78\%, respectively. Meanwhile, the radar charts in Figure \ref{fig-radar} underscore the remarkable robustness of AHEAD in maintaining high accuracy across diverse classes.
\subsubsection{Handling Data Sparsity}
\begin{figure*}[h]
  \centering
    \includegraphics[width=0.97\linewidth]{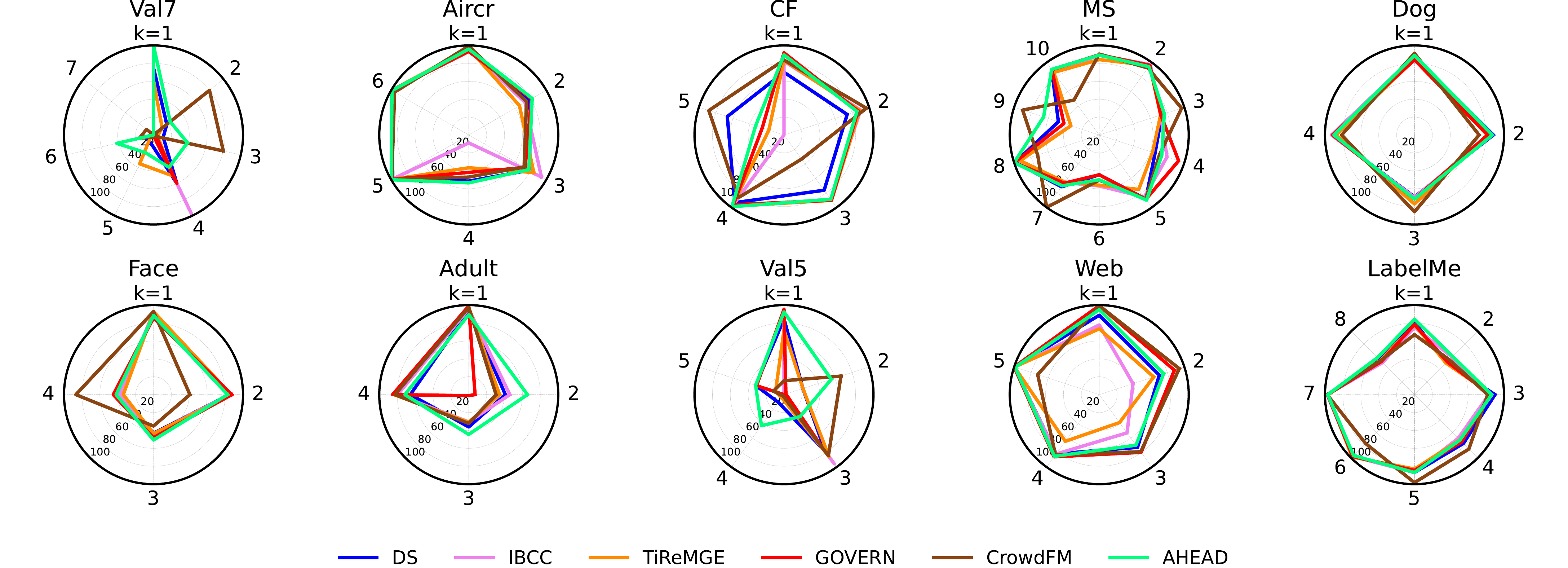}
    \caption{Class-wise accuracy (\%) comparison of AHEAD with representative confusion-matrix-based methods (DS, IBCC) and representation-learning-based methods (TiReMGE, GOVERN, CrowdFM).}
\label{fig-radar}
\end{figure*}

As shown in Table \ref{dataset}, the CF and Aircr datasets are extremely sparse, with median label counts ($Med_{w}$) of only 2 and 4, respectively, posing the trade-off between fine-grained annotator modeling and robust training. As shown in Table \ref{tab:performance}, confusion-matrix-based methods DS, IBCC, and CBCC suffer substantial performance degradation on these datasets. Specifically, DS ranks last on CF (79.67\%), while IBCC and CBCC rank last on Aircr (78.08\% and 76.56\%). However, they fail for different underlying reasons. The poor performance of DS on CF mainly stems from its simple annotator reliability estimation via maximum likelihood. In contrast, the degradation of IBCC and CBCC on Aircr is primarily due to their limited robustness to data sparsity and class imbalance: estimating a fine-grained $K \times K$ confusion matrix for an annotator with fewer than $K$ answered tasks is ill-posed, leading to overfitting and high variance. Notably, DS exhibits strong baseline generalization but ranks last on the CF dataset, highlighting the inherent difficulty of balancing models' expressive capacity with robustness to both data sparsity. By contrast, our AHEAD achieves top-tier accuracy across both datasets, as further evidenced in Figure \ref{fig-radar}-even when certain classes exhibit near-zero proportions.
\begin{figure*}[htbp]\label{val5-ms}
    \centering
    \begin{subfigure}[b]{0.44\textwidth}
        \centering
        \includegraphics[width=\textwidth]{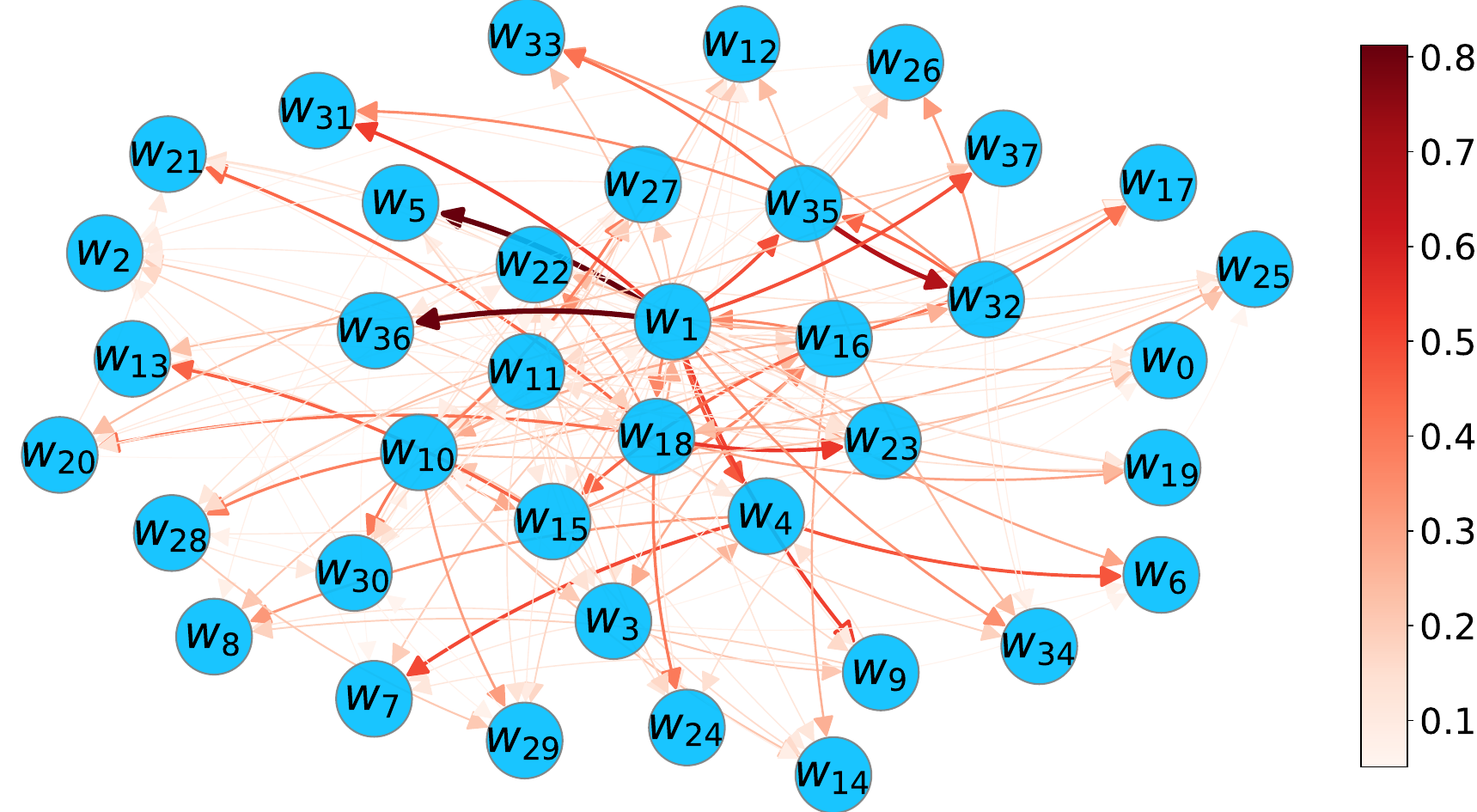} 
        \caption{Val5}
        \label{val5}
    \end{subfigure}
    \hspace{0.03\textwidth}
    \begin{subfigure}[b]{0.44\textwidth}
        \centering
        \includegraphics[width=\textwidth]{MS}
        \caption{MS}
        \label{ms}
    \end{subfigure}
    \caption{Cross-annotator correlations learned for Val5 and MS, where reliability propagates via the interaction topology.}
    \label{val5-ms}
\end{figure*}

Meanwhile, the representation-learning-based methods GOVERN, TiReMGE, and CrowdFM achieve competitive performance, suggesting the advantages of representation learning for capturing complex relationships. However, they are also sensitive to data sparsity; for instance, GOVERN performs worse on the Adult dataset, while TiReMGE shows inferior performance on the Web and Aircr datasets, which can be attributed to insufficient training data.

Overall, experimental results demonstrate that our cross-annotator learning effectively overcomes imprecise annotator estimation by leveraging the statistical strength of the entire annotator population rather than relying on individual data, yielding significantly better inference where prior methods fail.
\subsubsection{Handling Imbalanced Truth Distribution}
As presented in Table \ref{dataset}, the real-world datasets exhibit substantial diversity in class distributions, ranging from perfectly balanced scenarios (e.g., Face) to those characterized by severe imbalance (e.g., Val5, Val7, Adult, and Web). Notably, DS, IBCC, TiReMGE, and GOVERN display a marked sensitivity to class imbalance. This weakness manifests as a geometric collapse in Figure \ref{fig-radar}, particularly on datasets characterized by dominant or long-tail classes. The Val7 and Val5 datasets pose a dual challenge of severe class imbalance and extreme task difficulty, under which existing methods struggle significantly, with accuracies capped at 22\% and 41\%, respectively. Figure \ref{fig-radar} shows that on Val7, DS, IBCC, and TiReMGE collapse on four out of seven classes, and GOVERN collapses on six out of seven classes; while on Val5, DS, IBCC, and TiReMGE collapse on the majority class (class 4), and GOVERN collapses on the largest two classes (class 4 and class 2), resulting in highly skewed class-wise performance. In contrast, by explicitly modeling cross-annotator learning, AHEAD boosts accuracy on Val7 and Val5 to 36.90\% and 49.40\%, respectively, yielding improvements of 14.9\% and 8.4\%, respectively, highlighting AHEAD's robustness against class imbalance and strong capability in complex tasks.

Moreover, on the Adult dataset (where the majority class constitutes 56\%) and the Web dataset (where the minority and majority classes account for 12\% and 26\%, respectively), DS and IBCC tend to overfit the dominant classes (class 1 for Adult, and class 3 for Web), resulting in a skewed polygon shape that compromises the recall of other classes. Meanwhile, GOVERN collapses on minority classes (class 2 and class 3). By contrast, our AHEAD yields remarkable class-wise accuracy on these datasets, while preserving outstanding performance on balanced datasets like Face and Dog.

In summary, these experimental results demonstrate that AHEAD consistently maintains outstanding performance across classes.
\subsection{Case Studies on Val5 and MS}
Figure \ref{val5-ms} illustrates the cross-annotator attention coefficients that we learn from Val5 and MS, respectively.

Table \ref{dataset} reveals a distinct characteristic of Val5: the classes are imbalanced, and the number of tasks labeled per annotator follows an almost uniform distribution, with nearly every annotator labeling exactly 20 tasks.  Meanwhile, Table \ref{tab:performance} indicates that the annotations in this dataset are unreliable, as reflected by the consistently poor performance of all baselines, i.e., MV achieves an accuracy of only 35.17\%; although DS reaches 41\%, it collapses on the majority class (class 4). This scenario presents a particular challenge in distinguishing between the reliable and unreliable annotators. To address this, AHEAD's cross-annotator attention coefficients learned on Val5 form a dense topology, which facilitates reliable annotator estimation by effectively propagating information across the annotator population, as illustrated in Figure \ref{val5-ms}(a). For instance, although there is no direct connection between $w_{32}$ and $36$, their correlation is propagated via intermediate annotators $w_{1}$ and $w_{35}$. This mechanism enables AHEAD to effectively distinguish between reliable and unreliable annotators. As a result, AHEAD substantially improves the accuracy to 49.40\%.

\begin{figure*}[htbp]
    \centering
    \begin{subfigure}[b]{0.23\textwidth}
        \centering
        \includegraphics[width=\textwidth]{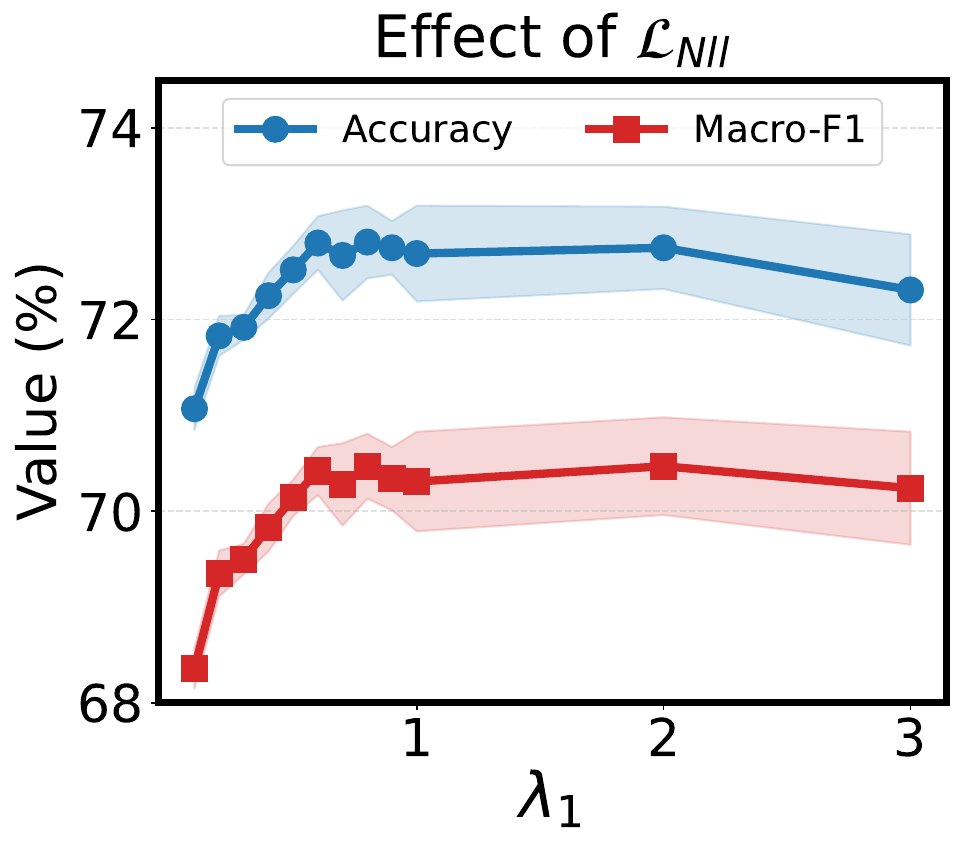} 
    \end{subfigure}
    \hfill 
    \begin{subfigure}[b]{0.23\textwidth}
        \centering
        \includegraphics[width=\textwidth]{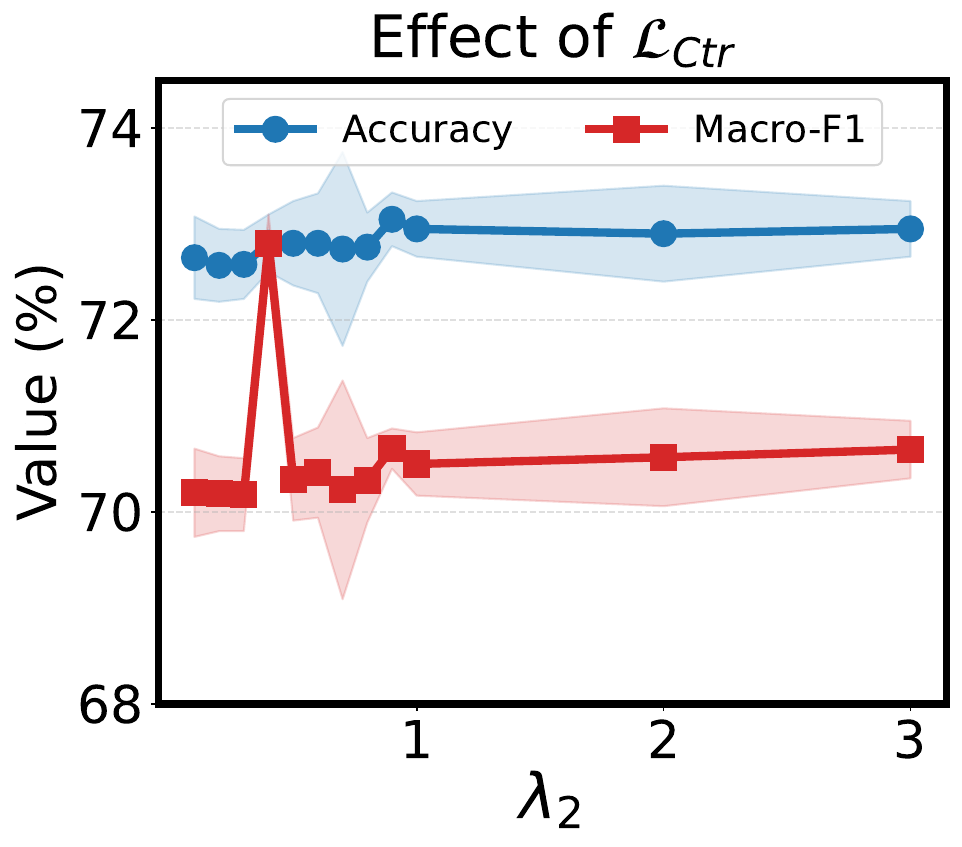}
    \end{subfigure}
    \hfill
    \begin{subfigure}[b]{0.23\textwidth}
    \centering
    \includegraphics[width=\textwidth]{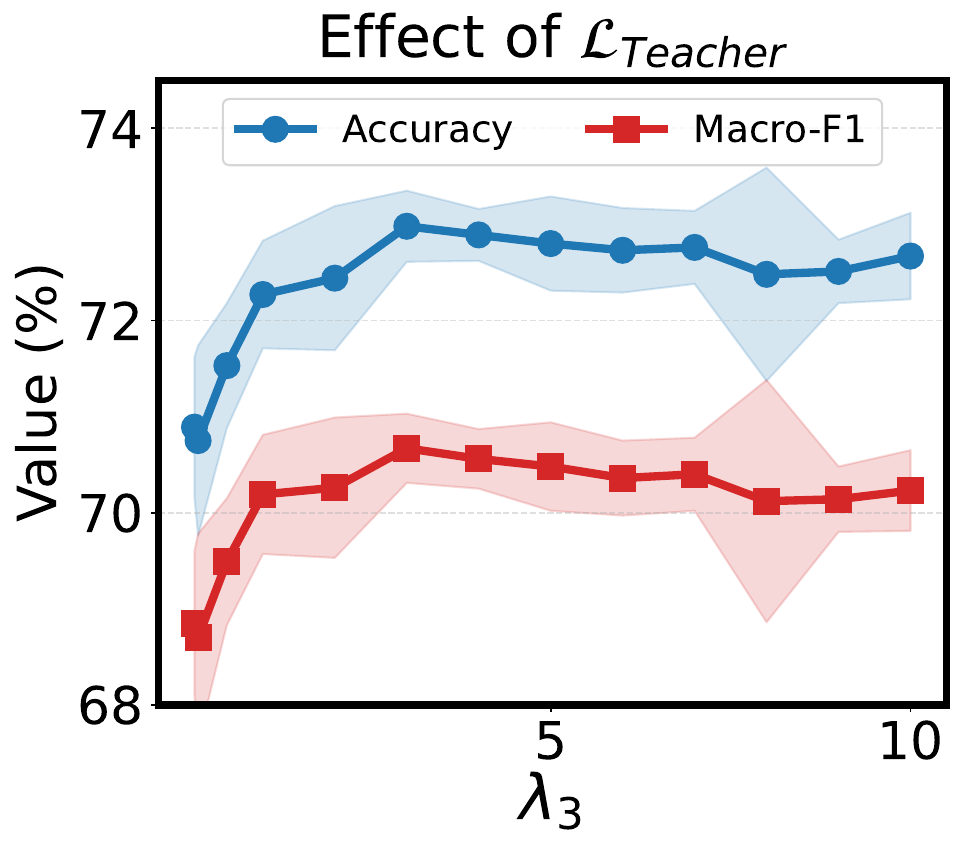}
    \end{subfigure}
    \hfill
    \begin{subfigure}[b]{0.23\textwidth}
        \centering
        \includegraphics[width=\textwidth]{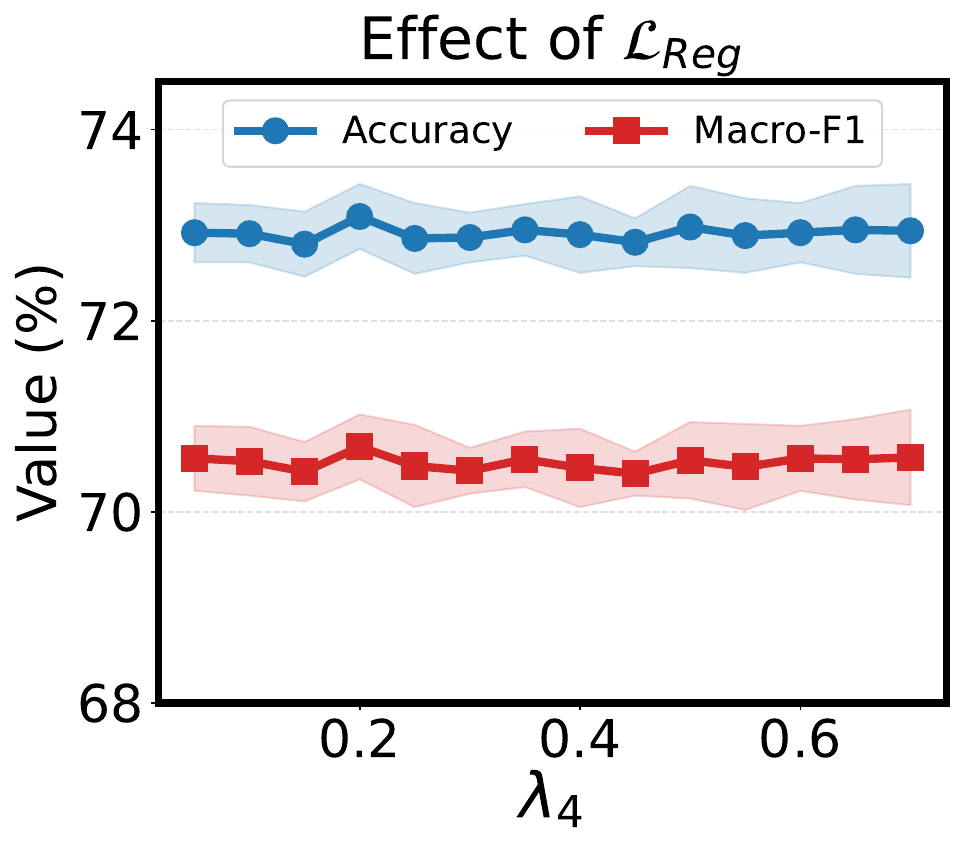}
    \end{subfigure}
    \caption{Effect of different $\lambda$ values on the model performance.}\label{ablation}
\end{figure*}
\begin{figure*}
\includegraphics[width=\textwidth]{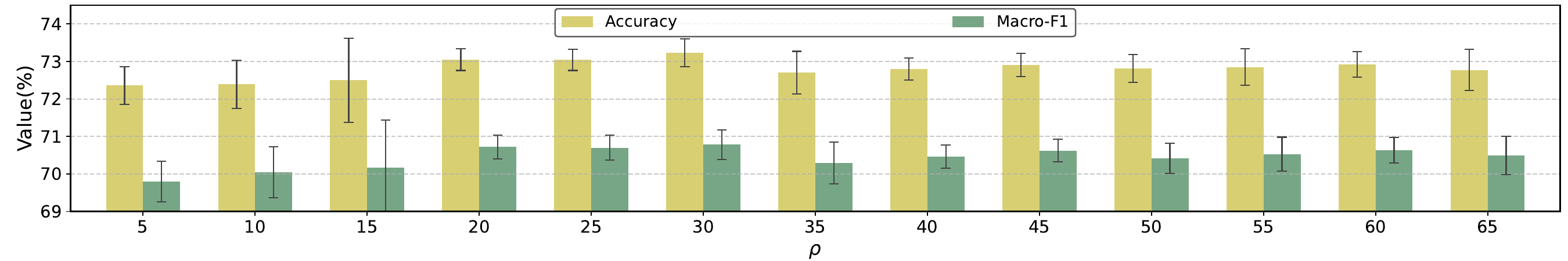}
\centering
\caption{Effect of different $\rho$ values on the model performance.} \label{percent-bar}
\end{figure*}
For the MS dataset, as shown in Table \ref{dataset}, while the class distribution is uniform, the number of tasks labeled per annotator follows a long-tail distribution: most annotators label only a small subset of tasks. This scenario poses the challenge of balancing model's complexity in discriminative capability with robustness to data sparsity. Table \ref{tab:performance} shows that the top-performing baselines are EBCC (78.71\%) and CATD (78.69\%). Meanwhile, representation-learning-based methods perform mediocrely due to the insufficient training data. As depicted in Figure \ref{val5-ms}(b), the cross-annotator attention coefficients learned by AHEAD reveal a deep, structured topology characterized by distinct clusters and stronger edge weights among specific nodes. By assigning high correlation weights to reliable sub-communities, AHEAD refines the aggregation process, achieving a state-of-the-art accuracy of 79.79\%.

The contrast between these topologies indicates that AHEAD is not limited to learn a static correlation pattern. Instead, it dynamically adapts its correlation structure, ranging from dense coupling to structured clustering to enable a robust performance.
\subsection{Ablation Studies}
\subsubsection{Effect of Loss Components}
To examine the contribution of each loss component to model performance, we perform a sensitivity analysis regarding the hyperparameters $\lambda_{1}$, $\lambda_{2}$, $\lambda_{3}$, and $\lambda_{4}$, which weigh the negative log-likelihood loss ($\mathcal{L}_{Nll}$), contrastive loss ($\mathcal{L}_{Ctr}$), teacher-student loss ($\mathcal{L}_{Teacher}$), and regularization loss ($\mathcal{L}_{Reg}$), respectively. As shown in Figure \ref{ablation}, we vary one hyperparameter at a time while fixing the others at their optimal values.

For the negative log-likelihood loss weight $\lambda_{1}$, the performance curve exhibits a clear inverted-V shape, peaking at $\lambda_{1}=1.0$; lower values lead to insufficient model training, whereas larger weights ($>1.0$) encourage overfitting to noisy pseudo-labels. The contrastive loss weight $\lambda_{2}$ has a relatively stable impact, with a local optimum around $\lambda_{2}=1.0$, suggesting that while feature alignment is beneficial, the model is robust to minor variations in this constraint. In contrast, the teacher-student loss weight $\lambda_{3}$ demonstrates a significant positive correlation with accuracy up to a threshold, reaching its optimal around $\lambda_{3}=6.0$. This large optimal value underscores the importance of leveraging supervisory signals from high-confidence annotators in mitigating bias and guiding optimization trajectory. Finally, the regularization loss weight $\lambda_{4}$ exhibits only minor variation within the range $[0.1, 0.7]$, suggesting that while constraining the confusion matrix is important for stability, its insensitive to precise tuning.

Overall, these findings confirm the necessity of each component, with the teacher-student loss $\mathcal{L}_{Teacher}$ from high-confidence annotators playing a dominant role in enhancing robustness. Meanwhile, the negative log-likelihood loss $\mathcal{L}_{Nll}$ ensures faithful fitting to observed labels, and the contrastive loss $\mathcal{L}_{Ctr}$ promotes meaningful feature alignment among annotators. In addition, the regularization loss $\mathcal{L}_{Reg}$ stabilizes the learning of confusion matrices and prevents degenerate solutions. Together, these components form a complementary objective that is critical to the effectiveness of AHEAD.
\begin{table*}[h]
\renewcommand{\arraystretch}{1.1}
\centering
\caption{Runtime (Sec) across different datasets and methods.}
\label{time-detail}
\begin{tabularx}{\textwidth}{lYYYYYYYYYYY}
\noalign{\hrule height 1pt}
Method & Val7 & Aircr & CF & MS & Dog & Face & Adult & Val5 & Web & LabelMe & Avg. \\
\noalign{\hrule height 1pt}
MV      & \textbf{0.005} & \textbf{0.003} & \textbf{0.01} & \textbf{0.004} & \textbf{0.005} & \textbf{0.004} & \textbf{0.05} & \textbf{0.004} & \textbf{0.01} & \textbf{0.04} & \textbf{0.01} \\
CATD    & 0.63 & 0.74 & 1.35 & 0.83 & 1.12 & 0.88 & 9.93 & 0.63 & 1.79 & 0.90 & 1.88 \\
GLAD    & 28.95 & 7.90 & 36.57 & 290.86 & 130.73 & 203.20 & 2561.32 & 19.06 & 506.23 & 40.74 & 382.56 \\
BWA     & \underline{0.05} & \underline{0.10} & \underline{0.02} & \underline{0.11} & \underline{0.05} & \underline{0.03} & \underline{0.39} & \underline{0.03} & \underline{0.10} & 0.20 & \underline{0.11} \\
\midrule
DS      & 0.98 & 1.09 & 0.78 & 1.02 & 1.07 & 0.88 & 7.79 & 0.82 & 1.96 & 2.35 & 1.87 \\
IBCC    & 3.45 & 3.85 & 4.08 & 7.31 & 7.02 & 5.47 & 66.05 & 3.15 & 14.05 & 5.78 & 12.02 \\
CBCC    & 5.31 & 5.44 & 13.04 & 9.39 & 6.24 & 4.88 & 32.01 & 4.49 & 9.99 & 7.59 & 9.84 \\
EBCC    & 2.10 & 9.99 & 18.56 & 35.37 & 4.66 & 8.99 & 371.59 & 1.44 & 65.95 & 15.16 & 53.38 \\
\midrule
TiReMGE & 46.74 & 55.15 & 74.31 & 82.57 & 127.84 & 93.78 & 1140.93 & 46.22 & 234.66 & 82.84 & 198.50 \\
GOVERN  & 10.15 & 15.54 & 15.31 & 20.49 & 43.51 & 30.91 & 1043.29 & 10.31 & 113.64 & 22.85 & 132.60 \\
CrowdFM  & 0.09 & 0.12 & 0.11 & 0.47 & 0.55 & 0.34 & 7.26 & 0.09 & 1.12 & \underline{0.16} & 1.03 \\
AHEAD   & 1.26 & 1.40 & 14.61 & 2.13 & 11.29 & 1.81 & 100.29 & 1.21 & 9.95 & 2.73 & 14.67 \\
\noalign{\hrule height 1pt}
\end{tabularx}
\end{table*}
\begin{table*}[h]\footnotesize
\renewcommand{\arraystretch}{1.25}
\centering
\caption{Comparison of accuracy (\%), Macro-F1 (\%), and runtime (Sec) on the largest dataset, Senti. OOM denotes an out-of-memory error on a GPU with 70GB of memory.}
\label{senti}
\begin{tabularx}{\textwidth}{lYYYYYYYYYYYc}
\noalign{\hrule height 1pt}
Metric & MV & CATD & GLAD &  BWA & DS &  IBCC & CBCC &  EBCC &  TiReMGE &       GOVERN & CrowdFM & AHEAD \\
\noalign{\hrule height 1pt}
\textit{Accu} \(\uparrow\)    & 88.26{\scriptsize$\pm$0.15}  &  88.30{\scriptsize$\pm$0.06} & \textbf{89.40{\scriptsize$\pm$0.00}} & 89.00{\scriptsize$\pm$0.00} &  82.40{\scriptsize$\pm$0.00}   & 85.90{\scriptsize$\pm$0.00} & 78.70{\scriptsize$\pm$0.00}  & 86.01{\scriptsize$\pm$0.14} &  90.35{\scriptsize$\pm$0.09} &   OOM  & 88.43{\scriptsize$\pm$0.77}  & 88.49{\scriptsize$\pm$0.21} \\
\noalign{\hrule height 1pt}
\textit{F1} \(\uparrow\)    & 78.10{\scriptsize$\pm$0.35}  &  78.71{\scriptsize$\pm$0.41} & \textbf{79.40{\scriptsize$\pm$0.00}} & 77.21{\scriptsize$\pm$0.00} &  73.70{\scriptsize$\pm$0.00}   & 75.94{\scriptsize$\pm$0.00} & 72.43{\scriptsize$\pm$0.00}  & 73.41{\scriptsize$\pm$0.48} &  78.99{\scriptsize$\pm$1.67} &   OOM  &  77.64{\scriptsize$\pm$1.11}  & 78.20{\scriptsize$\pm$0.46} \\
\noalign{\hrule height 1pt}
\textit{Time} \(\downarrow\)   & \textbf{0.227}  &  85.296 & 16721.323 & 3.224 &  76.664   & 35.833 & 239.631  & 10651.524 &  7988.357 &  OOM  & 37.560  & 1151.742 \\
\noalign{\hrule height 1pt}
\end{tabularx}
\end{table*}
\subsubsection{Effect of the Proportion of High-confidence Annotators.} Figure \ref{percent-bar} illustrates the effect of different $\rho$ values on the model performance. As shown in Figure \ref{percent-bar}, the model performance is relatively stable across different values of $\rho$. Both Accuracy and Macro-F1 improve as $\rho$ increases from small values, reach their best levels around $\rho=20~30$ and then remain largely unchanged with only minor fluctuations. This suggests that the model is not highly sensitive to $\rho$, and that a moderate value of $\rho$ is sufficient to achieve strong and robust performance.
\subsection{Efficiency}
Table \ref{time-detail} compares the runtime of various methods. Generally, the runtime of all methods increases with the number of labels. As table \ref{time-detail} turns out, MV consistently ranks first across datasets due to its simplicity, followed by BWA. The efficiency of BWA can be attributed to its single-parameter-based Bayesian formulation, which enables direct inference using the expectation-maximization framework. Notably, CrowdFM, which acquires transferable knowledge and can be directly deployed on new datasets, demonstrates runtime second only to MV and BWA. Meanwhile, as the simplest confusion matrix-based method, DS exhibits comparable performance to the single-parameter-based method CATD.

Meanwhile, while showing the clearly best \textit{Accuracy} and \textit{Macro-F1}, our AHEAD also exhibits competitive efficiency, e.g., it is orders of magnitude faster than GLAD, TiReMGE, and EBCC. As Table \ref{time-detail} shows, on the Adult dataset, AHEAD requires approximately $10^{2}$ seconds, whereas GLAD and TiReMGE exceed $10^{3}$ seconds. Furthermore, AHEAD maintains a runtime comparable to IBCC and CBCC, suggesting that it achieves an effective balance between model complexity and computational efficiency.

GLAD is the least efficient method, likely due to its explicit modeling of task difficulty and reliance on complex gradient-based optimization. Meanwhile, representation-learning-based methods TiReMGE and GOVERN incur substantial computational overhead due to complex network training and high-dimensional optimization of complex annotator-task dependencies, since the number of tasks can be several orders of magnitude larger than the number of annotators.

In general, methods that model annotators based on confusion matrices or representation learning require longer training time due to their elaborate parameterization and inference procedures, and our AHEAD achieves an effective balance between model complexity and computational efficiency.
\subsection{Scalability}
To further evaluate the scalability of all methods, we conduct experiments on the largest publicly available dataset, \textit{Senti}. Table \ref{senti} reports the performance results of all methods, demonstrating that GOVERN fails to run due to an out-of-memory error on a GPU with 70GB of memory, a limitation that has also been confirmed by its authors.

Table \ref{senti} demonstrates that our AHEAD achieves a strong balance between effectiveness and efficiency on Senti. Notably, although single-parameter-based methods such as MV and BWA are faster, AHEAD significantly outperforms them in predictive performance. At the same time, compared with more sophisticated probabilistic methods such as GLAD, EBCC, and TiReMGE, AHEAD achieves comparable or better performance with substantially lower runtime. This shows that AHEAD scales effectively to large-scale crowdsourcing data. Moreover, the foundation model CrowdFM also achieves outstanding performance due to its cross-dataset generalization mechanism, which can be directly deployed on new datasets.

Overall, the results demonstrate that AHEAD offers a desirable efficiency–effectiveness trade-off in large-scale settings, making it a practical and scalable choice for real-world crowdsourcing applications.
\section{Conclusion}
We identify a fundamental bottleneck in state-of-the-art multi-class label aggregation: unstable annotator reliability estimation incurred by data sparsity, i.e., most individual annotators only label a small subset of tasks. In response, we propose cross-annotator learning via representation learning and unify it with interpretable confusion matrices in an end-to-end framework. Specifically, we explicitly model cross-annotator learning over the extracted dense cross-annotator contexts, learning population-level annotator embeddings by aggregating individual-level annotator features with their contextual information. Then, we map these embeddings into interpretable confusion matrices to fit the observed labels. We formulate a composite objective that integrates likelihood maximization with structural regularization and directional guidance to facilitate effective training, where high-confidence annotators are proposed to alleviate the unsupervised training issues faced by existing models. Extensive experiments on 10 datasets with diverse characteristics demonstrate that AHEAD substantially improves the accuracy of multi-class label aggregation, achieving robust performance where prior methods fail. Meanwhile, the scalability experiments on the largest dataset demonstrate that AHEAD offers a desirable efficiency–effectiveness trade-off in large-scale settings.
\section*{AI-GENERATED CONTENT ACKNOWLEDGEMENT}
The authors used ChatGPT solely for proofreading and minor linguistic refinement. No part of the technical content, experimental design, or analysis was generated by ChatGPT or other AI tools.



\bibliographystyle{IEEEtran}
\bibliography{references}

@String{Computing = "Computing" }

@String{Computer = "{IEEE} Computer" }

@inproceedings{Li24,
  author = {Jiyi Li},
  year = {2024},
  title = {A comparative study on annotation quality of crowdsourcing and LLM via label aggregation},
  booktitle = {Proceedings of the International Conference on Acoustics, Speech and Signal Processing},
  pages = {6525--6529}
}

@inproceedings{Otanietal23,
  author  = {Mayu Otani and Riku Togashi and Yu Sawai and Ryosuke Ishigami and Yuta Nakashima and Esa Rahtu and Janne Heikkil{\"{a}} and Shin'ichi Satoh},
  year = {2023},
  title = {Toward verifiable and reproducible human evaluation for text-to-image generation},
  booktitle = {Proceedings of the {IEEE/CVF} Conference on Computer Vision and Pattern Recognition},
  pages = {14277--14286}
}

@inproceedings{Lietal2019a,
  author={Yuan Li and Benjamin I. P. Rubinstein and Trevor Cohn},
  year={2019},
  title={Exploiting worker correlation for label aggregation in crowdsourcing},
  booktitle={Proceedings of the International Conference on Machine Learning},
  pages={3886--3895}
}

@article{Chenetal25,
  author = {Ju Chen and Jun Feng and Shenyu Zhang and Xiaodong Li and Hamza Djigal},
  title = {Robust annotation aggregation in crowdsourcing via enhanced worker ability modeling},
  journal = {Information Processing \& Management},
  volume = {62},
  number = {1},
  pages = {103914},
  year = {2025}
}

@inproceedings{Lietal2019b,
  author={Yuan Li and Benjamin I. P. Rubinstein and Trevor Cohn},
  year={2019},
  title={Truth inference at scale: A bayesian model for adjudicating highly redundant crowd annotations},
  booktitle={Proceedings of the World Wide Web Conference},
  pages={1028--1038}
}

@article{DawidSkene1979,
  author={Alexander Philip Dawid and Allan M Skene},
  year={1979},
  title={Maximum likelihood estimation of observer error-rates using the {EM} algorithm},
  journal={Journal of the Royal Statistical Society. Series C (Applied Statistics)},
  volume={28},
  number={1},
  pages={20--28}
}

@inproceedings{KimGhahramani2012,
  author={Hyun-Chul Kim and Zoubin Ghahramani},
  year={2012},
  title={Bayesian classifier combination},
  booktitle={Proceedings of the International Conference on Artificial Intelligence and Statistics},
  pages={619--627}
}

@article{Wuetal2023a,
  author={Gongqing Wu and Xingrui Zhuo and Xianyu Bao and Xuegang Hu and Richang Hong and Xindong Wu},
  title={Crowdsourcing truth inference via reliability-driven multi-view graph embedding},
  journal={ACM Transactions on Knowledge Discovery from Data},
  volume={17},
  number={5},
  year={2023},
  pages={65:1--65:26}
}

@inproceedings{Venanzietal2014,
  author={Matteo Venanzi and John Guiver and Gabriella Kazai and Pushmeet Kohli and Milad Shokouhi},
  year={2014},
  title={Community-based bayesian aggregation models for crowdsourcing},
  booktitle={Proceedings of the World Wide Web Conference},
  pages={155--164}
}

@inproceedings{Caoetal2020,
  author={Ermei Cao and Difeng Wang and Jiacheng Huang and Wei Hu},
  year={2020},
  title={Open knowledge enrichment for long-tail entities},
  booktitle={Proceedings of the World Wide Web conference},
  pages={384--394}
}

@inproceedings{Whitehilletal2009,
  author={Jacob Whitehill and Paul Ruvolo and Tingfan Wu and Jacob Bergsma and Javier R. Movellan},
  year={2009},
  title={Whose vote should count more: Optimal integration of labels from labelers of unknown expertise},
  booktitle={Proceedings of the Annual Conference on Neural Information Processing Systems},
  pages={2035--2043}
}

@article{Lietal2014,
  author={Qi Li and Yaliang Li and Jing Gao and Lu Su and Bo Zhao and Murat Demirbas and Wei Fan and Jiawei Han},
  title={A confidence-aware approach for truth discovery on long-tail data},
  journal={Proceedings of the VLDB Endowment},
  volume={8},
  number={4},
  year={2014},
  pages={425--436}
}

@article{Songetal2021,
  author={Changyue Song and Kaibo Liu and Xi Zhang},
  title={Collusion detection and ground truth inference in crowdsourcing for labeling tasks},
  journal={Journal of Machine Learning Research},
  volume={22},
  number={190},
  year={2021},
  pages={8532-8576}
}

@article{Wuetal2023b,
  author = {Gongqing Wu and Xingrui Zhuo and Liangzhu Zhou and Xianyu Bao and Richang Hong and Xindong Wu},
  title = {{TIRA:} Truth Inference via Reliability Aggregation on Object-Source Graph},
  journal = {{IEEE} Transactions on Knowledge and Data Engineering},
  volume = {35},
  number = {11},
  pages = {11967--11981},
  year = {2023}
}

@inproceedings{Bonald17,
  author = {Thomas Bonald and Richard Combes},
  year = {2017},
  title = {A minimax optimal algorithm for crowdsourcing},
  booktitle = {Proceedings of the Annual Conference on Neural Information Processing Systems},
  pages = {4352--4360}
}

@inproceedings{Demartinietal2012,
  author={Gianluca Demartini and Djellel Eddine Difallah and Philippe Cudr{\'{e}}{-}Mauroux},
  title={Zencrowd: leveraging probabilistic reasoning and crowdsourcing techniques for large-scale entity linking},
  booktitle={Proceedings of the World Wide Web Conference},
  year={2012},
  pages={469--478}
}

@inproceedings{Ibrahietal23,
  author = {Shahana Ibrahim and Tri Nguyen and Xiao Fu},
  year = {2023},
  title = {Deep Learning From Crowdsourced Labels: Coupled Cross-Entropy Minimization, Identifiability, and Regularization},
  booktitle = {Proceedings of the International Conference on Learning Representations},
}

@inproceedings{Ibrahimetal19,
  author = {Shahana Ibrahim and Xiao Fu and Nikolaos Kargas and Kejun Huang},
  year = {2019},
  title = {Crowdsourcing via pairwise co-occurrences: identifiability and algorithms},
  booktitle = {Proceedings of the Annual Conference on Neural Information Processing Systems},
  pages = {7845--7855},
}

@inproceedings{Chu21b,
  author = {Zhendong Chu and Jing Ma and Hongning Wang},
  year = {2021b},
  title = {Learning from crowds by modeling common confusions},
  booktitle = {Proceedings of the {AAAI} Conference on Artificial Intelligence},
  pages = {5832--5840}
}

@inproceedings{Ibrahim21,
  author = {Shahana Ibrahim and Xiao Fu},
  year = {2021},
  title = {Crowdsourcing via annotator co-occurrence imputation and provable symmetric nonnegative matrix factorization},
  booktitle = {Proceedings of the International Conference on Machine Learning},
  pages = {4544--4554}
}

@inproceedings{Zhangetal24,
  author = {Hansong Zhang and Shikun Li and Dan Zeng and Chenggang Yan and Shiming Ge},
  year = {2024},
  title = {Coupled confusion correction: Learning from crowds with sparse annotations},
  booktitle = {Proceedings of the {AAAI} Conference on Artificial Intelligence},
  pages = {16732--16740}
}

@article{Zhangetal16,
  author = {Yuchen Zhang and Xi Chen and Dengyong Zhou and Michael I. Jordan},
  title = {Spectral methods meet {EM:} {A} provably optimal algorithm for crowdsourcing},
  journal = {Journal of Machine Learning Research},
  volume = {17},
  year = {2016},
  pages = {102:1--102:44}
}

@inproceedings{Hornetal18,
  author = {Grant Van Horn and Steve Branson and Scott Loarie and Serge J. Belongie and Pietro Perona},
  year = {2018},
  title = {Lean Multiclass Crowdsourcing},
  booktitle = {Proceedings of the {IEEE} Conference on Computer Vision and Pattern Recognition},
  pages = {2714--2723}
}

@inproceedings{Aydinetal14,
  author = {Bahadir Ismail Aydin and Yavuz Selim Yilmaz and Yaliang Li and Qi Li and Jing Gao and Murat Demirbas},
  year = {2014},
  title = {Crowdsourcing for Multiple-Choice Question Answering},
  booktitle = {Proceedings of the {AAAI} Conference on Artificial Intelligence},
  pages = {2946--2953}
}

@inproceedings{Liuetal24,
  author = {Hao Liu and Jiacheng Liu and Feilong Tang and Peng Li and Long Chen and Jiadi Yu and Yanmin Zhu and Min Gao and Yanqin Yang and Xiaofeng Hou},
  year = {2024},
  title = {Graph Contrastive Learning for Truth Inference},
  booktitle = {Proceedings of the International Conference on Data Engineering},
  pages = {263--275}
}

@inproceedings{liu2026crowdfm,
    author={Hao Liu and Jiacheng Liu and Feilong Tang and Long Chen and Jiadi Yu and Yanmin Zhu and Qiwen Dong and Yichuan Yu and Xiaofeng Hou},
    year={2026},
    title={Towards a Foundation Model for Crowdsourced Label Aggregation},
    booktitle={Proceedings of the International Conference on Learning Representations}
}

@inproceedings{Yinetal17,
  author = {Li'ang Yin and Jianhua Han and Weinan Zhang and Yong Yu},
  year = {2017},
  title = {Aggregating Crowd Wisdoms with Label-aware Autoencoders},
  booktitle = {Proceedings of the International Joint Conference on Artificial Intelligence},
  pages = {1325--1331}
}

@article{Lyuetal21,
  author = {Shanshan Lyu and Wentao Ouyang and Yongqing Wang and Huawei Shen and Xueqi Cheng},
  title = {Truth Discovery by Claim and Source Embedding},
  journal = {{IEEE} Transactions on Knowledge and Data Engineering},
  volume = {33},
  number = {3},
  year = {2021},
  pages = {1264--1275}
}

@inproceedings{Chu21a,
  author = {Zhendong Chu and Hongning Wang},
  year = {2021a},
  title = {Improve Learning from Crowds via Generative Augmentation},
  booktitle = {Proceedings of the {ACM} {SIGKDD} Conference on Knowledge Discovery and Data Mining},
  pages = {167--175}
}

@inproceedings{Yangetal24,
  author = {Boyi Yang and Liangxiao Jiang and Wenjun Zhang},
  year = {2024},
  title = {Probabilistic Matrix Factorization-based Three-stage Label Completion for Crowdsourcing},
  booktitle = {Proceedings of the {IEEE} International Conference on Data Mining},
  pages = {540--549}
}

@article{Zhangetal25a,
  author = {Wenjun Zhang and Liangxiao Jiang and Chaoqun Li},
  title = {{ELDP:} Enhanced Label Distribution Propagation for Crowdsourcing},
  journal = {{IEEE} Transactions on Pattern Analysis and Machine Intelligence},
  volume = {47},
  number = {3},
  pages = {1850--1862},
  year = {2025}
}

@inproceedings{Yangetal23,
  author = {Yonghui Yang and Zhengwei Wu and Le Wu and Kun Zhang and Richang Hong and Zhiqiang Zhang and Jun Zhou and Meng Wang},
  year = {2023},
  title = {Generative-Contrastive Graph Learning for Recommendation},
  booktitle = {Proceedings of the International {ACM} {SIGIR} Conference on Research and Development in Information Retrieval},
  pages = {1117--1126}
}

@article{Liuetal25,
  author={Jiacheng Liu and Feilong Tang and Hao Liu and Long Chen and Yichuan Yu and Yanmin Zhu and Jiadi Yu and Xiaofeng Hou and Pheng-Ann Heng},
  title={BAT: A Versatile Bipartite Attention-Based Approach for Comprehensive Truth Inference in Mobile Crowdsourcing},
  journal={IEEE Transactions on Mobile Computing},
  volume={24},
  number={10},
  pages={9368-9382},
  year={2025}
  }

@inproceedings{Liuetal21,
  author={Jiacheng Liu and Feilong Tang and Jielong Huang},
  year={2021},
  title={Truth Inference with Bipartite Attention Graph Neural Network from a Comprehensive View},
  booktitle={Proceedings of the IEEE International Conference on Multimedia and Expo},
  pages={1-6}
}

@article{Das2023state,
  author={Anubrata Das and Houjiang Liu and Venelin Kovatchev and Matthew Lease},
  title={The state of human-centered {NLP} technology for fact-checking},
  journal={Information Processing \& Management},
  volume={60},
  number={2},
  pages={103219},
  year={2023}
}

@article{modaresnezhad2020,
  author={Minoo Modaresnezhad and Lakshmi S. Iyer and Prashant Palvia and Vasyl Taras},
  title={Information Technology {(IT)} enabled crowdsourcing: A conceptual framework},
  journal={Information Processing \& Management},
  volume={57},
  number={2},
  pages={102--135},
  year={2020}
}

@article{li2020,
  author={Yuanbing Li and Xian Wu and Yifei Jin and Jian Li and Guoliang Li},
  title={Efficient algorithms for crowd-aided categorization},
  journal={Proceedings of the VLDB Endowment},
  volume={13},
  number={8},
  pages={1221--1233},
  year={2020}
}

@article{Zhengetal2017,
  author={Yudian Zheng and Guoliang Li and Yuanbing Li and Caihua Shan and Reynold Cheng},
  title={Truth inference in crowdsourcing: Is the problem solved?},
  journal={Proceedings of the VLDB Endowment},
  volume={10},
  number={5},
  pages={541--552},
  year={2017}
}

@article{Lietal2016,
  author={Guoliang Li and Jiannan Wang and Yudian Zheng and Michael J. Franklin},
  title={Crowdsourced data management: A survey},
  journal={IEEE Transactions on Knowledge and Data Engineering},
  volume={28},
  number={9},
  pages={2296--2319},
  year={2016}
}

@inproceedings{Lietal14,
  author={Qi Li and Yaliang Li and Jing Gao and Bo Zhao and Wei Fan and Jiawei Han},
  year={2014},
  title= {Resolving conflicts in heterogeneous data by truth discovery and source reliability estimation},
  booktitle={Proceedings of the International Conference on Management of Data},
  pages={1187--1198}
}

@inproceedings{Zhouetal12,
  author={Dengyong Zhou and John C. Platt and Sumit Basu and Yi Mao},
  year={2012},
  title={Learning from the Wisdom of Crowds by Minimax Entropy},
  booktitle={Proceedings of the Annual Conference on Neural Information Processing Systems},
  pages={2204--2212}
}

@inproceedings{Zhengetal22,
  author = {Libin Zheng and Peng Cheng and Lei Chen and Jianxing Yu and Xuemin Lin and Jian Yin},
  year = {2022},
  title = {Crowdsourced Fact Validation for Knowledge Bases},
  booktitle = {Proceedings of the International Conference on Data Engineering},
  pages = {938--950}
}

@inproceedings{Chaietal19,
  author = {Chengliang Chai and Ju Fan and Guoliang Li and Jiannan Wang and Yudian Zheng},
  year = {2019},
  title = {Crowdsourcing Database Systems: Overview and Challenges},
  booktitle = {Proceedings of the International Conference on Data Engineering},
  pages = {2052--2055}
}

@inproceedings{Sunetal24,
  author = {Yushi Sun and Jiachuan Wang and Peng Cheng and Libin Zheng and Lei Chen and Jian Yin},
  year = {2024},
  title = {Cross-Domain-Aware Worker Selection with Training for Crowdsourced Annotation},
  booktitle = {Proceedings of the International Conference on Data Engineering},
  pages = {249--262}
}

@article{Zhangetal224,
  author = {Zhang, Chen Jason and Yunrui Liu and Pengcheng Zeng and Ting Wu and Lei Chen and Pan Hui and Fei Hao},
  title = {Similarity-driven and task-driven models for diversity of opinion in crowdsourcing markets},
  journal = {The VLDB Journal},
  volume = {33},
  number = {5},
  pages = {1377--1398},
  year = {2024}
}

@inproceedings{Deetal25,
  author = {Soham De and Michiel A. Bakker and Jay Baxter and Martin Saveski},
  year = {2025},
  title = {Supernotes: Driving Consensus in Crowd-Sourced Fact-Checking},
  booktitle = {Proceedings of the World Wide Web Conference},
  pages = {3751--3761}
}

@inproceedings{Gienappetal25,
  author = {Lukas Gienapp and Tim Hagen and Maik Fr{\"{o}}be and Matthias Hagen and Benno Stein and Martin Potthast and Harrisen Scells},
  year = {2025},
  title = {The Viability of Crowdsourcing for {RAG} Evaluation},
  booktitle = {Proceedings of the International {ACM} {SIGIR} Conference on Research and Development in Information Retrieval},
  pages = {159--169}
}

@inproceedings{Qiuetal24,
  author = {Jielin Qiu and Andrea Madotto and Zhaojiang Lin and Paul A. Crook and Yifan Ethan Xu and Babak Damavandi and Xin Dong and Christos Faloutsos and Lei Li and Seungwhan Moon},
  year = {2024},
  title = {SnapNTell: Enhancing Entity-Centric Visual Question Answering with Retrieval Augmented Multimodal {LLM}},
  booktitle = {Proceedings of the findings of the Association for Computational Linguistics},
  pages = {247--266}
}

@article{RodriguesPR13,
  author = {Filipe Rodrigues and Francisco C. Pereira and Bernardete Ribeiro},
  title = {Learning from multiple annotators: Distinguishing good from random labelers},
  journal = {Pattern Recognition Letters},
  volume = {34},
  number = {12},
  pages = {1428--1436},
  year = {2013}
}

@book{Durrett2010,
  author = {Rick Durrett},
  title = {Probability: Theory and Examples, 4th Edition},
  publisher = {Cambridge University Press},
  year = {2010}
}

@inproceedings{TannoSSAS19,
  author = {Ryutaro Tanno and Ardavan Saeedi and Swami Sankaranarayanan and Daniel C. Alexander and Nathan Silberman},
  year = {2019},
  title = {Learning From Noisy Labels by Regularized Estimation of Annotator Confusion},
  booktitle = {Proceedings of the findings of the {IEEE} Conference on Computer Vision and Pattern Recognition},
  pages = {11244--11253}
}

@inproceedings{Lietal21,
  author = {Xuefeng Li and Tongliang Liu and Bo Han and Gang Niu and Masashi Sugiyama},
  year = {2021},
  title = {Provably End-to-end Label-noise Learning without Anchor Points},
  booktitle = {Proceedings of the International Conference on Machine Learning},
  pages = {6403--6413}
}

@inproceedings{IbrahimN023,
  author = {Shahana Ibrahim and Tri Nguyen and Xiao Fu},
  year = {2023},
  title = {Deep Learning From Crowdsourced Labels: Coupled Cross-Entropy Minimization, Identifiability, and Regularization},
  booktitle = {Proceedings of the International Conference on Learning Representations}
}

@inproceedings{Tanetal24,
  author = {Zhi Qin Tan and Olga Isupova and Gustavo Carneiro and Xiatian Zhu and Yunpeng Li},
  year = {2024},
  title = {Bayesian Detector Combination for Object Detection with Crowdsourced Annotations},
  booktitle = {Proceedings of the European Conference on Computer Vision},
  pages = {329--346}
}

@article{Zhangetal25b,
  author = {Yue Zhang and Yiyi Chen and Chaowei Fang and Qian Wang and Jiayi Wu and Jingmin Xin},
  title = {Learning from open-set noisy labels based on multi-prototype modeling},
  journal = {Pattern Recognition},
  volume = {157},
  pages  = {110902},
  year = {2025}
}

@article{Wuetal22,
  author = {Haiqin Wu and Liangmin Wang and Ke Cheng and Dejun Yang and Jian Tang and Guoliang Xue},
  title = {Privacy-Enhanced and Practical Truth Discovery in Two-Server Mobile Crowdsensing},
  journal = {IEEE Transactions on Network Science and Engineering},
  volume = {9},
  number = {3},
  pages = {1740--1755},
  year = {2022}
}

@article{Baietal25,
  author = {Jing Bai and Jinsong Gui and Tian Wang and Houbing Song and Anfeng Liu and Neal N. Xiong},
  title = {{ETBP-TD:} An Efficient and Trusted Bilateral Privacy-Preserving Truth Discovery Scheme for Mobile Crowdsensing},
  journal = {{IEEE} Transactions on Mobile Computing},
  volume = {24},
  number = {3},
  pages = {2203--2219},
  year = {2025}
}

@ArtifactSoftware{R,
    title = {R: A Language and Environment for Statistical Computing},
    author = {{R Core Team}},
    organization = {R Foundation for Statistical Computing},
    address = {Vienna, Austria},
    year = {2019},
    url = {https://www.R-project.org/},
}


\end{document}